\pdfoutput=1

\documentclass[11pt]{article}

\usepackage[final]{acl}
\usepackage{tikz}

\usepackage{times}
\usepackage{latexsym}
\usepackage{amssymb}
\usepackage{amsmath}
\usepackage{algorithm}  
\usepackage{algpseudocode}

\usepackage{subcaption} 
\usepackage{caption} 

\usepackage{array}
\usepackage{multirow} 
\usepackage{booktabs} 

\usepackage[T1]{fontenc}

\usepackage[utf8]{inputenc}

\usepackage{silence}
\usepackage{microtype}

\usepackage{inconsolata}

\usepackage{graphicx}

\title{KSHSeek: Data-Driven Approaches to Mitigating and Detecting Knowledge-Shortcut Hallucinations in Generative Models}

 \author{First Author \\
   Affiliation / Address line 1 \\
   Affiliation / Address line 2 \\
  Affiliation / Address line 3 \\
  \texttt{email@domain} \\\And
  Second Author \\
   Affiliation / Address line 1 \\
   Affiliation / Address line 2 \\
   Affiliation / Address line 3 \\
   \texttt{email@domain} \\}

\author{
  \textbf{Zhongxin Liu\textsuperscript{1}},
  \textbf{Zhiwei Wang\textsuperscript{2}},
  \textbf{Jun Niu\textsuperscript{1}},
  \textbf{Ying Li\textsuperscript{2}},
  \textbf{Hongyu Sun\textsuperscript{3}},
\\
  \textbf{Meng Xu\textsuperscript{4}},
  \textbf{He Wang\textsuperscript{1}},
  \textbf{Gaofei Wu\textsuperscript{1}},
  \textbf{Yuqing Zhang\textsuperscript{1,2,3}}
\\
\\
  \textsuperscript{1}Xidian University,
  \textsuperscript{2}University of Chinese Academy of Sciences,
  \textsuperscript{3}Hainan University,
\\
  \textsuperscript{4}University of Waterloo
\\
 \small{
  \href{mailto:email@domain}{liuzhongxin@stu.xidian.edu.cn}}
  }

\begin{document}
\maketitle
\begin{abstract}

The emergence of large language models (LLMs) has significantly advanced the development of natural language processing (NLP), especially in text generation tasks like question answering. However, model hallucinations remain a major challenge in natural language generation (NLG) tasks due to their complex causes. We systematically expand on the causes of factual hallucinations from the perspective of knowledge shortcuts, analyzing hallucinations arising from correct and defect-free data and demonstrating that knowledge-shortcut hallucinations are prevalent in generative models. To mitigate this issue, we propose a \emph{high similarity pruning algorithm} at the data preprocessing level to reduce spurious correlations in the data. Additionally, we design a specific detection method for knowledge-shortcut hallucinations to evaluate the effectiveness of our mitigation strategy. Experimental results show that our approach effectively reduces knowledge-shortcut hallucinations, particularly in fine-tuning tasks, without negatively impacting model performance in question answering. This work introduces a new paradigm for mitigating specific hallucination issues in generative models, enhancing their robustness and reliability in real-world applications.
\end{abstract}


\section{Introduction}

The emergence of large language models (LLMs)
has brought a paradigm shift to natural language processing (NLP),
especially in generative tasks such as question-answering
(\citealp{rangapur2024battle, michail2023uzh_clyp, qin2023chatgpt})
%
However,
this revolution has also caused a growing concern,
known as model hallucinations.
Huang et al.\cite{Huang:23} building on the definition of hallucinations proposed by Ji et al.(\citealp{NPH-2021-neural}; \citealp{ji2023survey}), expanded the applicability and scope of the term, classifying model hallucinations into two types: factual hallucinations and faithfulness hallucinations. This expanded classification provides a new paradigm for understanding model hallucinations.

We focus on factual hallucinations, and have observed a critical fact: training data has played a significant role in causing factual hallucinations. One notable example is the ``floating-point comparison hallucination''\footnote{\url{https://x.com/goodside/status/1812977352085020680}},
%
%
When the prompt \emph{"9.11 or 9.9, which number is larger?"} is given to
LLMs,
many existing commercial LLMs provide incorrect answers, as illustrated in Appendix A, Table \ref{tab:counts}.
%
\begin{figure}[t]
  \includegraphics[width=\linewidth]{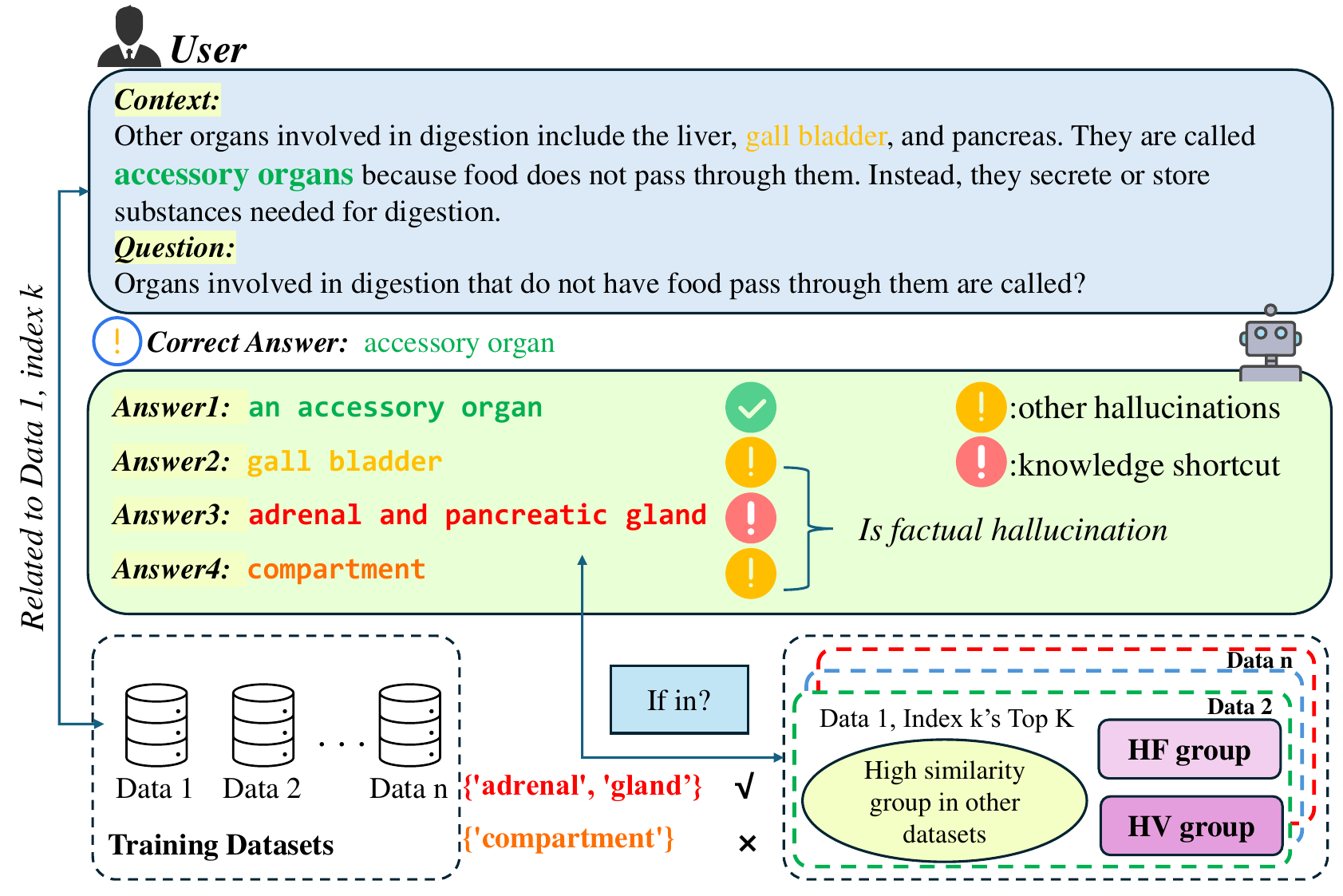}
  \caption{An example of what is the knowledge-shortcut hallucinations in CQA tasks}
  \label{fig:example of KSH}
\end{figure}

\begin{table*}[htbp!]
    \centering
    \resizebox{\textwidth}{!}{ 
    \begin{tabular}{lllllll}
    \toprule
    \textbf{Index} & \textbf{C-Q} & \textbf{Correct-Answer} & \textbf{Generate-Answer} & \textbf{Jaccard-sim} & \textbf{TF\_IDF-sim} & \textbf{AI-sim} \\
    \toprule
    before 1 & <|user|>Other...& accessory organs & \colorbox{yellow}{liver} & 0.00000 & 0.00000 & 0.39682 \\
    before 2 & <|user|>Other...& accessory organs & \colorbox{yellow}{assistant} \colorbox{green}{organ} & 0.00000 & 0.00000 & \textbf{0.54355} \\
    before 3 & <|user|>Other...& accessory organs & \colorbox{red}{adrenal} and \colorbox{orange}{pancreatic} \colorbox{red}{gland} & 0.00000 & 0.00000 & 0.30573 \\
    before 4 & <|user|>Other...& accessory organs & \colorbox{orange}{compartment} & 0.00000 & 0.00000 & 0.46903 \\\hline
    after 1 & <|user|>Other...& accessory organs & an \colorbox{green}{accessory organ} & \textbf{1.00000} & \textbf{0.70930} & \textbf{0.90820}\\
    after 2 &<|user|>Other... & accessory organs & \colorbox{yellow}{liver} & 0.00000 & 0.00000 & 0.39682 \\
    after 3 &<|user|>Other... & accessory organs & \colorbox{green}{accessory organ} & \textbf{1.00000} & \textbf{1.00000} & \textbf{0.94474} \\
    after 4 &<|user|>Other... & accessory organs & \colorbox{orange}{attached} \colorbox{green}{organ} & \textbf{0.33333} & \textbf{0.33610} & \textbf{0.74186} \\
    \toprule
    \end{tabular}
    }
    \caption{Examples of CQA tasks before and after mitigation. \colorbox{green}{Green}: Correct words that appear in the C-Q text. \colorbox{yellow}{Yellow}: Incorrect words that appear in the C-Q text but are not the correct answer. \colorbox{orange}{Orange}: Incorrect words that do not appear in the C-Q text, representing other hallucinations. \colorbox{red}{Red}: Words appearing in the high-frequency and high-value groups, indicating knowledge-shortcut hallucinations. See Appendix A for detailed information}
    \label{tab:example before and after mitigation}
\end{table*}

A major cause of the aforementioned hallucinations is knowledge-shortcut(\citealp{ju2024investigating};\citealp{li2022pre}).
%
The training data often contains a significant amount of information such as computer system version numbers and book indexes. LLMs
have learned the comparative features of this data and erroneously applied these features to the comparison of regular numbers, leading to hallucinations. In the classification proposed by Li\cite{li2022pre}, this cause is referred to as a "Knowledge Shortcut". Building on this concept, we define hallucinations caused by knowledge shortcuts as knowledge-shortcut hallucinations.

Knowledge shortcut arises because language models typically do not genuinely understand the intricate and complex factual knowledge but rather rely on shortcuts. They tend to over-rely on semantically proximate positions in the pre-training data, shared high-frequency words, and the quantity of related documents \cite{kandpal23}.This can introduce spurious correlation biases, causing the model to produce hallucinations even when working with correct and defect-free data sources.
%
%

We focus on factual hallucinations and introduce a Context-Question-Answer (CQA) task to analyze hallucinations caused by knowledge shortcuts, termed knowledge-shortcut hallucinations. In a CQA task, the correct answer typically resides in the context, and answers from large models that deviate from the correct answer are considered factual hallucinations. However, not all factual hallucinations are knowledge-shortcut hallucinations. When the model's answer is not in the context but is found in the high-similarity group of the CQ (shown in red in Figure \ref{fig:example of KSH}), it is considered a knowledge-shortcut hallucination. In contrast, answers found in the context (like \textbf{\emph{Answer2(yellow)}}) or outside both the context and high-similarity group (like \textbf{\emph{Answer4(orange)}}) are called other hallucinations.

A common approach for mitigating data-related hallucinations is data filtering, including strictly controlling data source(\citealp{gao2020pile}; \cite{gunasekar2023textbooks}) and deduplication. Deduplication which is divided into exact and near duplicates faces challenges. Exact duplicate detection is inefficient for large datasets \cite{manber1993suffix}, while near duplicate method like hash-based algorithm MinHash \cite{broder1997minhash} prioritize speed but miss hidden information. Semantic duplicate recognition using pre-trained models \cite{abbas2023semdedup} is slower and impractical for large datasets. Thus, balancing granularity in duplicate detection and processing speed, while effectively reducing knowledge-shortcut hallucinations, remains a challenge.

This paper focuses on analyzing knowledge-shortcut hallucinations triggered by high-similarity texts from correct, defect-free data. We propose a \emph{High Similarity Pruning Algorithm} that mitigates knowledge-shortcut hallucinations from a data perspective by leveraging semantic similarity, shared high-frequency words. Furthermore, based on these characteristics and incorporating model uncertainty(\citealp{xiao2021hallucination}; \citealp{miao2023selfcheck}), we design a hybrid detection method tailored for CQA tasks to identify knowledge-shortcut hallucinations effectively. Our mitigation strategy demonstrates promising results across multiple LLMs and parameter scales. Notably, in the fine-tuning of nanoGPT-large, it successfully reduces knowledge-shortcut hallucinations by 6.5\%.

Table \ref{tab:example before and after mitigation} shows the same color-coding as in Figure \ref{fig:example of KSH}. We present a real CQA example and compare responses before and after mitigation. Detailed results in appendix \ref{example} show significant improvement across similarity metrics, a reduction in knowledge-shortcut hallucinations, and overall higher response quality.

Overall, the contributions of our paper can be summarized as follows:

\begin{itemize}
\item We investigate the mechanisms and patterns underlying knowledge-shortcut hallucinations driven by accurate and defect-free data. We identify their general characteristics and reveal their widespread presence in LLMs.

\item We propose a novel detection method that combines semantic similarity and the uncertainty of LLM-generated outputs in CQA tasks. This method enables the quantitative evaluation of knowledge-shortcut hallucinations across different LLMs and training strategies (e.g., fine-tuning vs. training from scratch).

\item To mitigate knowledge-shortcut hallucinations, we introduce a \emph{Data High Similarity Pruning Algorithm} based on the identified generation mechanisms of such hallucinations. Quantitative evaluations demonstrate that this algorithm significantly improves the generation quality of LLMs and excels in suppressing hallucinations. The source code for our approach is available at github.
\end{itemize}


\section{Methodology}

\begin{figure*}[t]
  \includegraphics[width=\linewidth]{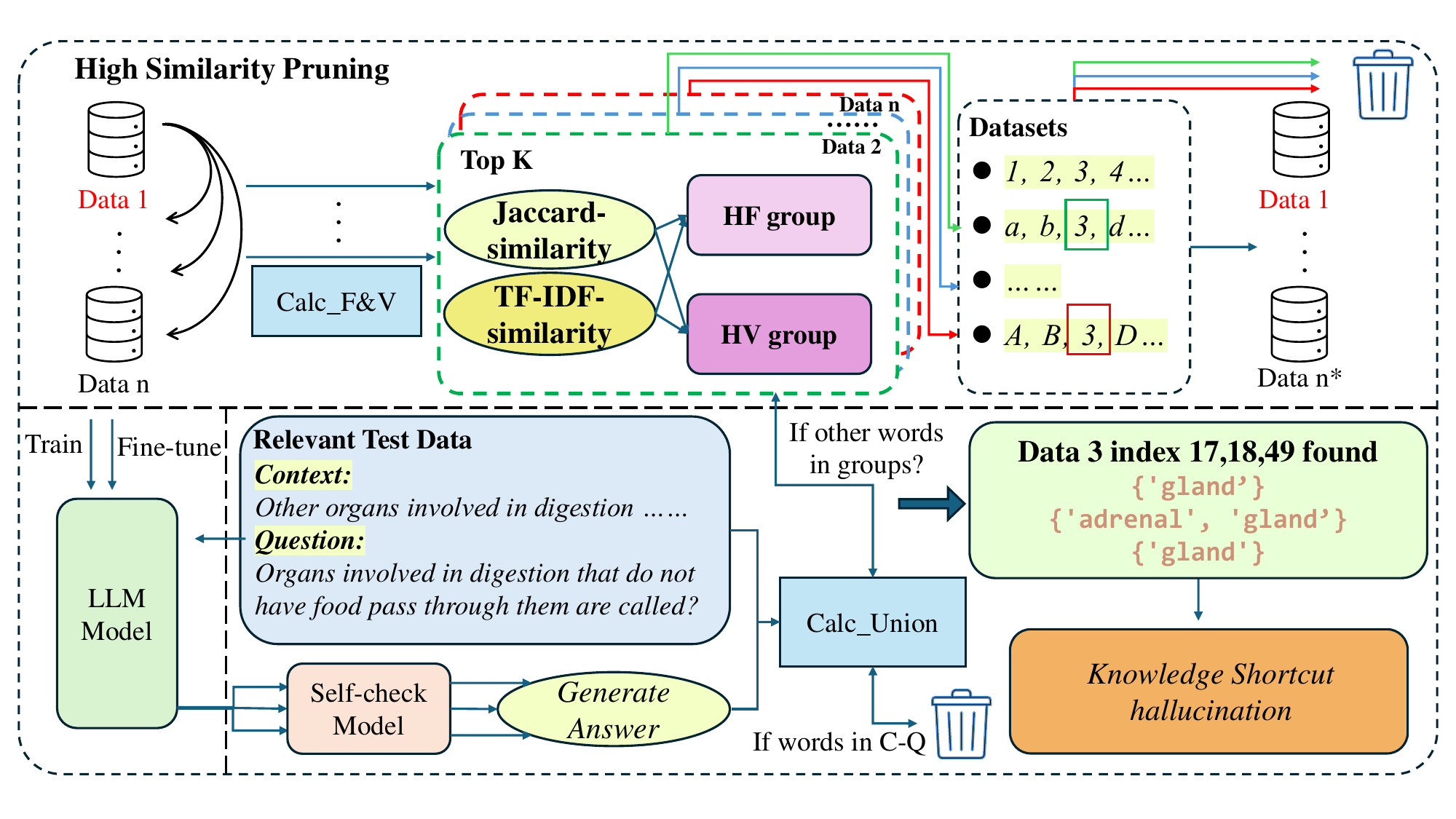}
  \caption {Overview of detection and mitigation.}
  \label{fig:structure}
\end{figure*}

\subsection{Overview}
The CQA task, characterized by simple answers and clear facts, is particularly well-suited for the study of knowledge-shortcut hallucinations. Through prior analysis, we found that such hallucinations arise from the misleading effect of high-frequency co-occurring and highly similar words in the training data (e.g., “9.11” being interpreted as a computer version number or directory, which is larger than “9.9”). The essence of our mitigation strategy is to filter high-frequency co-occurring and highly similar terms in the training data based on specific metrics (reducing or balancing the occurrences of "9.11" as a version number or directory). The focus of knowledge-shortcut hallucinations detection is to determine whether a factual error is caused by the misdirection of high-frequency and highly similar co-occurring entries in the training data, based on the background and the question (e.g., whether a comparison of computer version numbers or directories, like "9.11 is larger than 9.8," exists in the training data). 

Building on these insights, we have refined our knowledge-shortcut mitigation strategy (Section \ref{text:mitigation}) and hallucination detection approach (Section  \ref{text:detection}), with the overall framework shown in Figure \ref{fig:structure}. 

We carefully selected three metrics for measuring text similarity: Jaccard similarity, TF-IDF similarity, and pre-trained model similarity. Through extensive engineering optimization, we ensured that our text similarity metrics not only maintain fine granularity but also enhance runtime efficiency, as detailed in Appendix \ref{metrics of similarity}. Experimental results (Section \ref{experiments}, Appendix \ref{results}) confirm that our mitigation strategy is simple to implement and highly effective, demonstrating the robustness of the strategy and validating its performance in knowledge-shortcut hallucination detection.

\subsection{Data High Similarity Pruning}
\label{text:mitigation}
We define high-frequency co-occurring words and highly similar words as the high similarity group, with the specific concept outlined in Appendix \ref{HS group}. Based on the definition of the high-similarity group, we designed the \emph{High Similarity Pruning Algorithm} shown in Figure \ref{fig:structure}, which helps generative models reduce the occurrence of knowledge-shortcut hallucinations.

Given a batch of fine-tuning or training data from $n$ different categories $(data_1, data_2, ..., data_n)$, we define the following steps for \emph{data1}: with hyperparameters $(K_1, K_2, \alpha_1, \alpha_2)$, we compute the set $R_{1,j\in n}$ for deletion:

\textbf{1)} For each row in $data_1$, compute the top $K_1$ Jaccard and TF-IDF similarity values with the remaining $(n-1)$ datasets. Record the corresponding indices and values.  
\textbf{2)} Calculate the top $K_2$ most frequent indices (High-Frequency group, $G_{HF}$) and the top $K_2$ largest values (High-Value group, $G_{HV}$).  
\textbf{3)} Combine the four groups (\emph{HF} and \emph{HV} for both Jaccard and TF-IDF similarities), remove duplicates, and identify rows in $data_{j\in n}$ to delete, denoted as $R_{1,j\in n}$(Equation \ref{get R}):  
\begin{equation}
  \label{get R}
  R_{1,j\in n}=Set(\alpha_1G_{HF}+\alpha_2G_{HV})
\end{equation}
\textbf{4)} Iterate over all $n$ datasets to compute the final set $R_{all}$ for deletion across all datasets with Equation \ref{get R all}:  
\begin{equation}
  \label{get R all}
  R_{all}=Set(\sum_{i=1}^n\sum_{j\neq i}^nR_{i,j})
\end{equation}

\subsection{Detection of Knowledge-Shortcut Hallucination}
\label{text:detection}
Detecting knowledge-shortcut hallucinations requires distinguishing them from other hallucinations. We focus on fact-based question-answering tasks with a CQA structure, where the correct answer is embedded in the context.
To detect knowledge-shortcut hallucinations, we propose a method combining similarity features and self-check uncertainty measurement\cite{miao2023selfcheck}. The pseudocode is in Appendix \ref{sec:appendix pseudocode}.

\textbf{1)} For a given context-question pair ($CQ$), we compute the most similar entry $CQA_{ij}$ from the datasets $(data_1, ..., data_n)$, where $i$ denotes the dataset and $j$ the row index.  
\textbf{2)} We calculate the Jaccard and TF-IDF similarity scores between this entry and others from $data_{k\neq i}$, identifying the high-frequency group ($G_{HF}$) and high-value group ($G_{HV}$).  
\textbf{3)} In the self-check module, the model generates an answer $A_o$. We then regenerate $m$ responses $(A_1, \dots, A_m)$ from the same input. If $A_o$ significantly differs from $(A_1, \dots, A_m)$, the response is flagged as a potential hallucination. The variation is quantified by Equation \ref{measure difference}, where $m=5$ and $\alpha_3=0.2$.  
\begin{equation}
  \label{measure difference}
  \frac{\sum_{l=1}^m 1-Sim(A_o,A_{l})}{m}\leq \alpha_3
\end{equation}
\textbf{4)} We compute the difference set $S_o$ between $A_o$ and $CQ$ (Equation \ref{difference}). If $S_o$ is non-empty, the process continues:  
\begin{equation}
  \label{difference}
  S_o=Set(A_o) - Set(CQ)
\end{equation}
\textbf{5)} Finally, we calculate the intersection between $S_o$ and the high-frequency ($G_{HF}$) and high-value ($G_{HV}$) groups from $CQA$ (Equation \ref{caculate intersection}). If non-empty, we conclude that $A_o$ is a knowledge-shortcut hallucination.  
\begin{equation}
  \label{caculate intersection}
  Set(A_o)\cap Set(CQA_{G_{HF},G_{HV}})
\end{equation}

\subsection{Metrics of Effectiveness Evaluation}
To evaluate the mitigation method, we design metrics that measure performance differences of the same model under identical parameter configurations, both before and after applying the method.

\textbf{Coarse-grained metrics}: 1) Number of Non-Zero and Non-Empty Similarity Rows: Count the rows where the similarity between generated and correct answers is non-zero and non-empty, before and after mitigation. 2) Average Similarity: Calculate the average similarity across the test set. These metrics offer a macroscopic view of the method's overall impact.


\textbf{Fine-grained metrics}: The fine-grained approach directly counts the number of knowledge-shortcut hallucinations in the test set. By comparing the numbers before and after applying the mitigation method, this metric offers a straightforward and clear evaluation of the method’s effectiveness.

\section{Experiments}
\label{experiments}
\subsection{Experiments Setting}
\subsubsection{Datasets}
We selected four datasets with a CQA structure from the generative text datasets on Hugging Face as training or fine-tuning datasets $(data_1, data_2, data_3, data_4)$, along with a hallucination test dataset as the ablation test dataset. The four CQA datasets belong to different domains, forming a diverse training or fine-tuning datasets. In terms of data quantity selection, not all data from the four datasets were used. The large discrepancies in the total volume of data across different datasets could make it difficult for the model to learn long-tail knowledge\cite{kandpal23}, thus negatively impacting the experimental results. Therefore, we selected portions of data from \emph{trivia-cqa\footnote{\url{https://huggingface.co/datasets/tilyupo/trivia_cqa}}} and \emph{QASports\footnote{\url{https://huggingface.co/datasets/PedroCJardim/QASports}}} that closely matches the sample sizes of \emph{sciq\footnote{\url{https://huggingface.co/datasets/allenai/sciq}}} and \emph{financial-qa-10K\footnote{\url{https://huggingface.co/datasets/virattt/financial-qa-10K}}}. Details of this data selection can be found in Table \ref{tab:datasets}.
\begin{table}
  \centering
  \resizebox{\linewidth}{!}{
  \begin{tabular}{lcc}
    \hline
    \textbf{Dataset} & \textbf{Category} & \textbf{Number of rows}\\
    \hline
    sciq    & science &  11679    \\
    financial-qa-10K     & finance &  7000     \\
    trivia-cqa     & miscellaneous &  14000    \\
    QASports     & Basketball &  14453    \\
    \hline
  \end{tabular}}
  \caption{General description of the CQA datasets}
  \label{tab:datasets}
\end{table}

For the test sets, we selected 600 samples from the \emph{sciq} test dataset, a natural sciences dataset focused on objective facts, as the related test set. Additionally, we selected 513 samples from the \emph{llm hallucination\footnote{\url{https://huggingface.co/datasets/C0uchP0tat0/llm_hallucinations}}} test dataset as an unrelated test set to evaluate the method's performance under different conditions.

\subsubsection{Model Selection}
We conducted our experiments on two generative models which is distributed under the MIT License. We used the model according to the terms specified in the license: nanoGPT\footnote{\url{https://github.com/karpathy/nanoGPT}} and TinyLlama\footnote{\url{https://github.com/jzhang38/TinyLlama}}. For nanoGPT, we selected three parameter scales: gpt2-large (774M), gpt2-medium (350M), and gpt2 (124M), to perform fine-tuning and training experiments. For TinyLlama\cite{zhang2024tinyllama}, we conducted fine-tuning experiments using the LoRA\cite{hulora} (Low-Rank Adaptation) method at the 1.1B parameter scale.

\subsubsection{Implementation Details}
Our experiments consist of three phases: assessment, mitigation, and detection.

\textbf{Assessment:} We used the \emph{sciq} dataset ($data_1$) and progressively combined it with three additional datasets to form four training datasets. We trained three nanoGPT models (gpt2-large, gpt2-medium, gpt2) using both training and fine-tuning approaches, while the TinyLlama model was fine-tuned only. All models were evaluated on the \emph{sciq test} using the CQA task. The results from the models trained on $sciq(data_1)$ alone served as the baseline for the related test set. We assessed the similarity changes when additional datasets were incorporated, analyzing both increases and decreases. Since there is no baseline for the unrelated test set, similarity changes are not evaluated for it.

\textbf{Mitigation:} We compared the performance of the models before and after applying the mitigation method, using consistent test sets. The mitigation parameters were set as: $(K_1, K_2, \alpha_1, \alpha_2) = (50, \text{lens} \times 0.006, 0.4, 0.1)$. $K_1 = 50$ corresponds to the Top-K parameter in nanoGPT, a key factor in hallucination generation. $K_2$ is related to dataset length, with a value of 0.006 to avoid excessive data removal. The value of $0.4$ prioritizes high-frequency overlapping data in pruning. We chose $\alpha_1 + \alpha_2 = 0.5$ to balance the influence of High-Frequency (HF) and High-Value (HV) groups. The \emph{High Similarity Pruning} increases data source independence, reducing semantic overlap between unrelated categories.
\begin{figure*}[t]

  \includegraphics[width=0.3\linewidth]{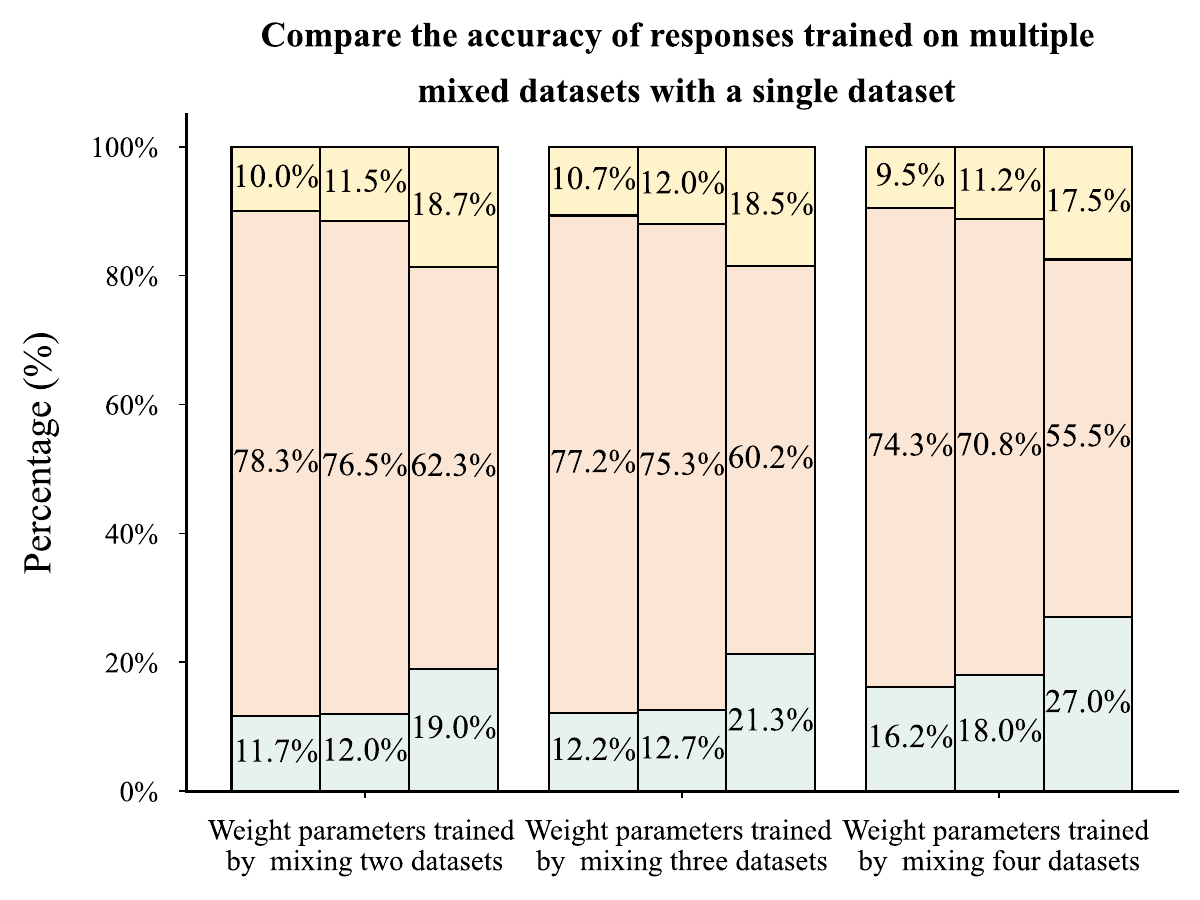} \hfill
  \includegraphics[width=0.3\linewidth]{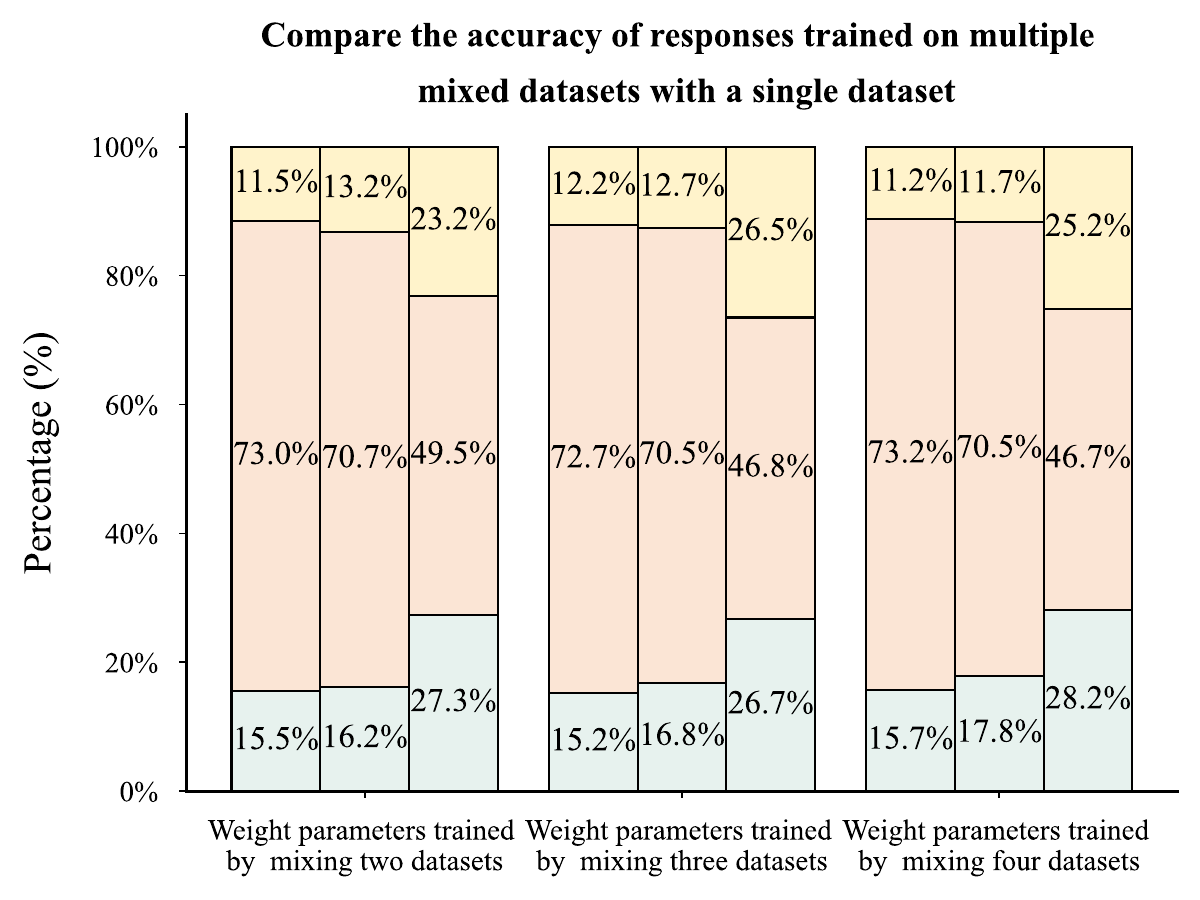} \hfill
  \includegraphics[width=0.36\linewidth]{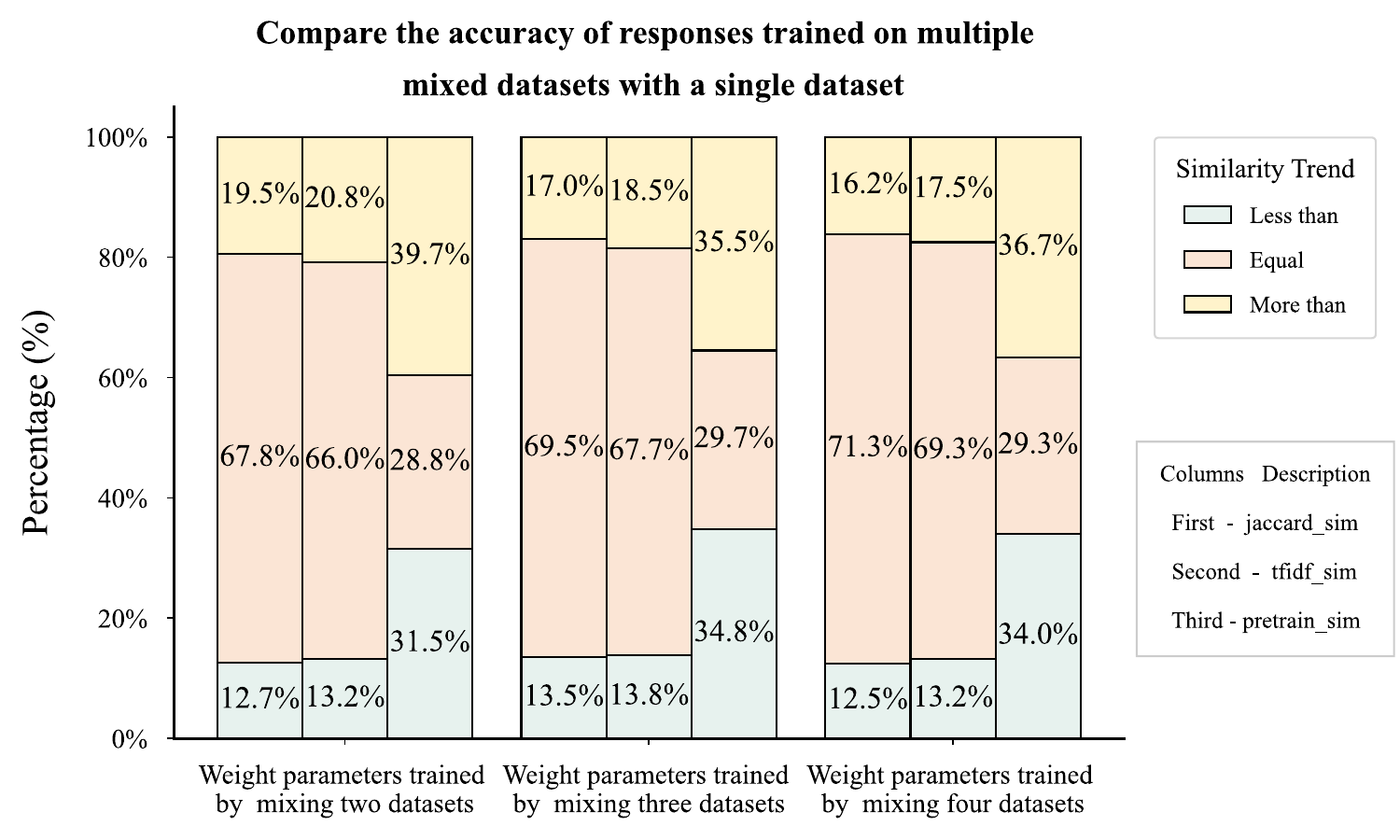}
  \caption {Evaluation of nanoGPT models under the fine-tuning method across three parameter scales: normal, medium and large(from left to right). Each x-axis represents a different similarity metric: Jaccard similarity, TF-IDF similarity, and Pre-trained model-based similarity, respectively.}
  \label{fig:ft-before}
\end{figure*}
\begin{figure*}[t]

  \includegraphics[width=0.3\linewidth]{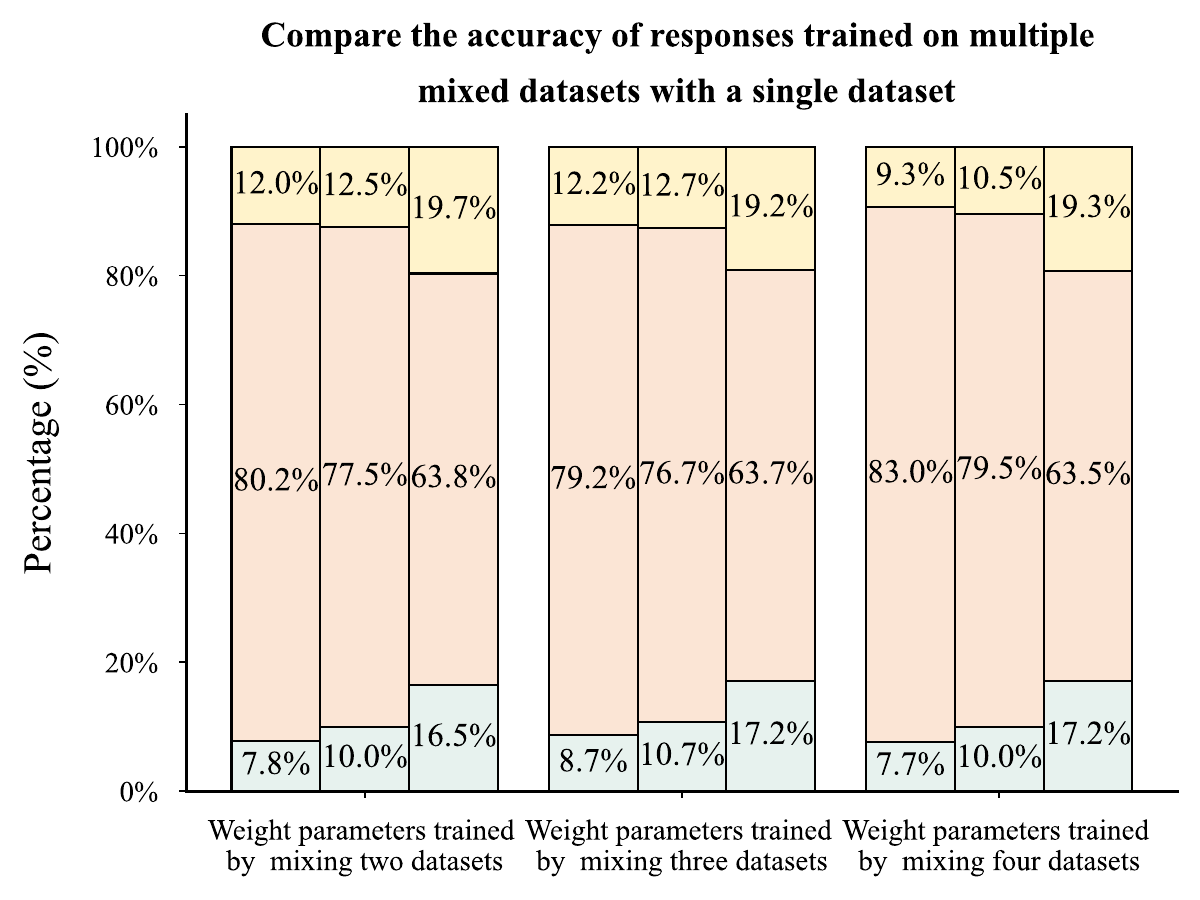} \hfill
  \includegraphics[width=0.3\linewidth]{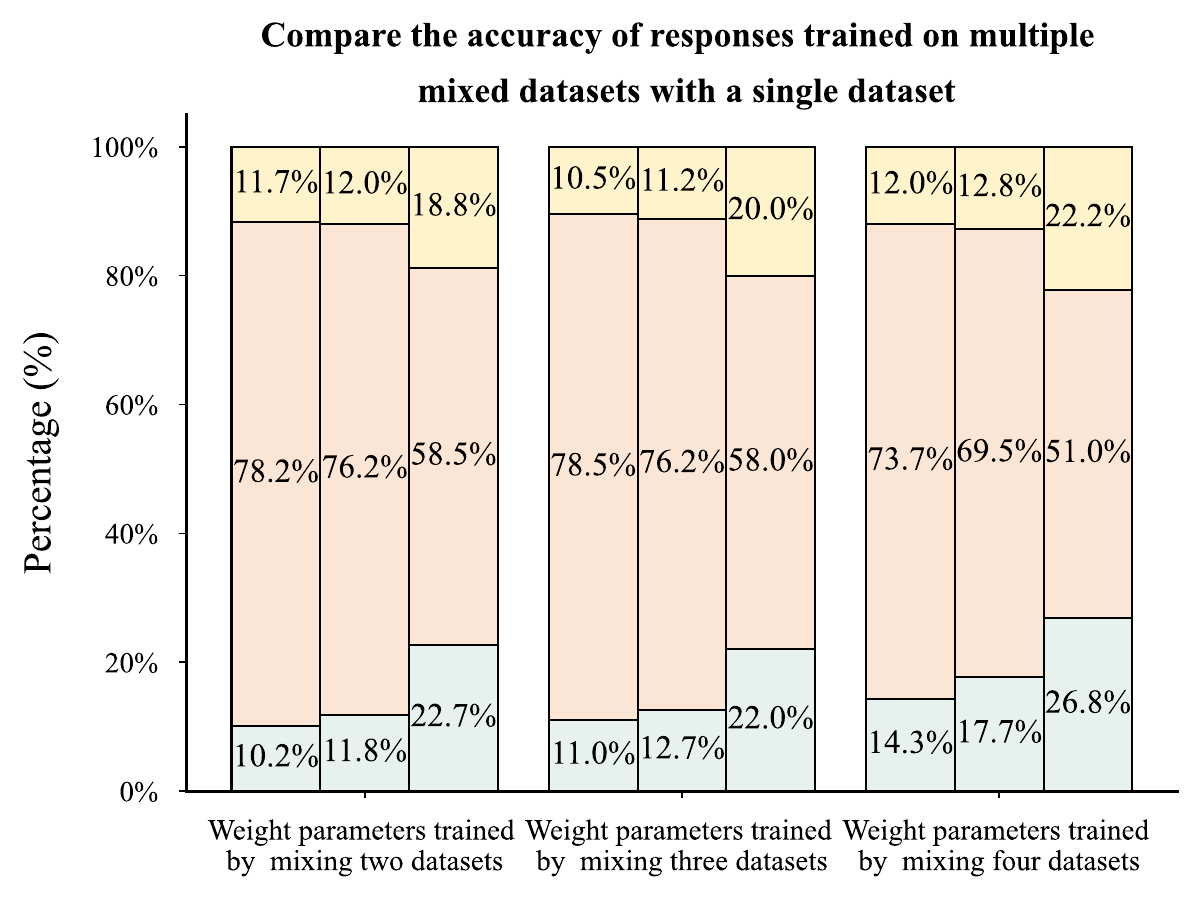} \hfill
  \includegraphics[width=0.36\linewidth]{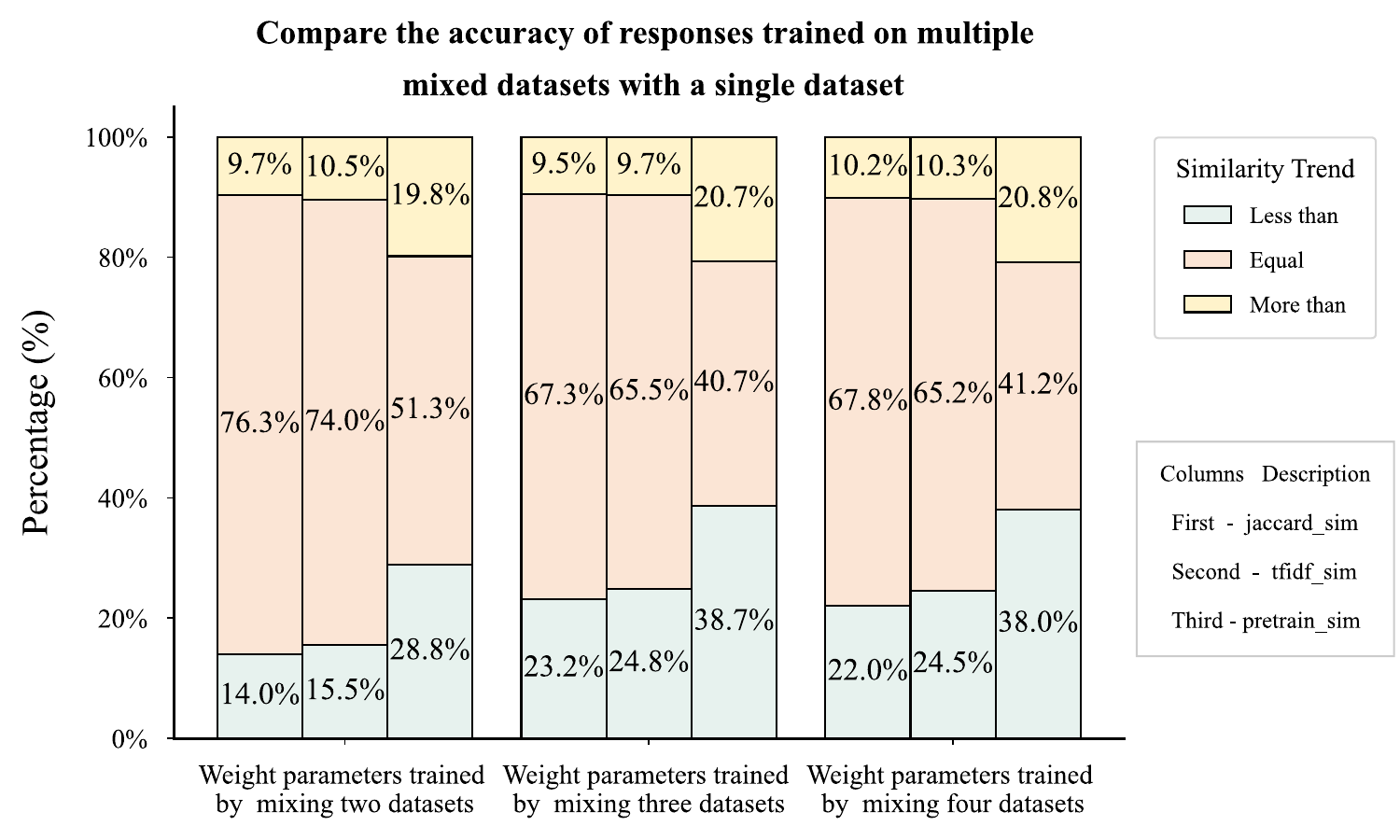}
  \caption {Evaluation of nanoGPT models under the training method across three parameter scales: normal, medium and large(from left to right). Each x-axis represents a different similarity metric: Jaccard similarity, TF-IDF similarity, and Pre-trained model-based similarity, respectively.}
  \label{fig:train-before}
\end{figure*}

For instance, applying the \emph{data high similarity pruning algorithm} to the \emph{sciq test} yields the updated data quantities, as shown in Table \ref{tab:after mitigation}.

\begin{table}
  \centering
  \resizebox{\linewidth}{!}{
  \begin{tabular}{lcc}
    \hline
    \textbf{Dataset}  & \textbf{Number of rows} & \textbf{Reduction magnitude}\\
    \hline
    sciq     &  11679  & 0\%  \\
    financial-qa-10K     &  6962 & 0.542\%   \\
    trivia-cqa     &  13926  &  0.529\%  \\
    QASports      &  14376  &  0.533\% \\
    \hline
  \end{tabular}}
  \caption{For sciq test, after mitigation}
  \label{tab:after mitigation}
\end{table}

\textbf{Detection:} We performed detection and metric evaluations for all models before and after mitigation on both test sets. To ensure result stability, we repeated experiments with different random seeds. The experiments were conducted on two NVIDIA 3090 GPUs (24 GB each). The source code is available on GitHub.

\subsection{Experiment 1: Evaluation—-The Prevalence of Knowledge-Shortcut Hallucinations}
We trained models on the four datasets described in Table 1 and conducted CQA tasks on a related test set comprising 600 samples. Using the results from training on $sciq(data_1)$ as the baseline, we evaluated the accuracy changes in model answers to the same questions after mixing additional datasets into the training data.  

Models trained on a single category of data and tested on the corresponding category’s test set can fully demonstrate the model's performance. By progressively mixing other datasets into the training data and repeating the testing process, we revealed the widespread presence of knowledge-shortcut hallucinations from a macro perspective.  

As shown in Figure \ref{fig:ft-before}, for models fine-tuned with different parameter scales, the similarity between generated answers and correct answers, measured by all three similarity metrics (Jaccard similarity, TF-IDF similarity, and pre-trained model similarity), decreased to varying extents as more datasets were mixed in. Compared to the baseline, models trained with multiple mixed datasets showed a higher proportion of "less" labels than "more" labels in their responses (The model trained on the \emph{sciq} dataset serves as the baseline. If the responses generated by models trained with additional mixed datasets show an increase in any of the three similarity metrics, they are labeled as "more"; if the similarity decreases, they are labeled as "less"). Similar trends were observed in models trained using the full training method, as illustrated in Figure \ref{fig:train-before}. The evaluation results and analysis after applying the mitigation strategy can be found in Appendix \ref{after mitigation result}.  

This phenomenon is also observed in the fine-tuning of the \emph{TinyLlama 1.1B} model, with the corresponding results presented in Appendix \ref{tinyllama result}.

Our evaluations demonstrate that knowledge-shortcut hallucinations are widespread across various datasets and models. This trend is consistently observed across different model architectures, parameter scales, and training methods, highlighting the significant impact of dataset composition on the accuracy and reliability of generative models.
\begin{figure*}[t]
  \includegraphics[width=0.48\linewidth]{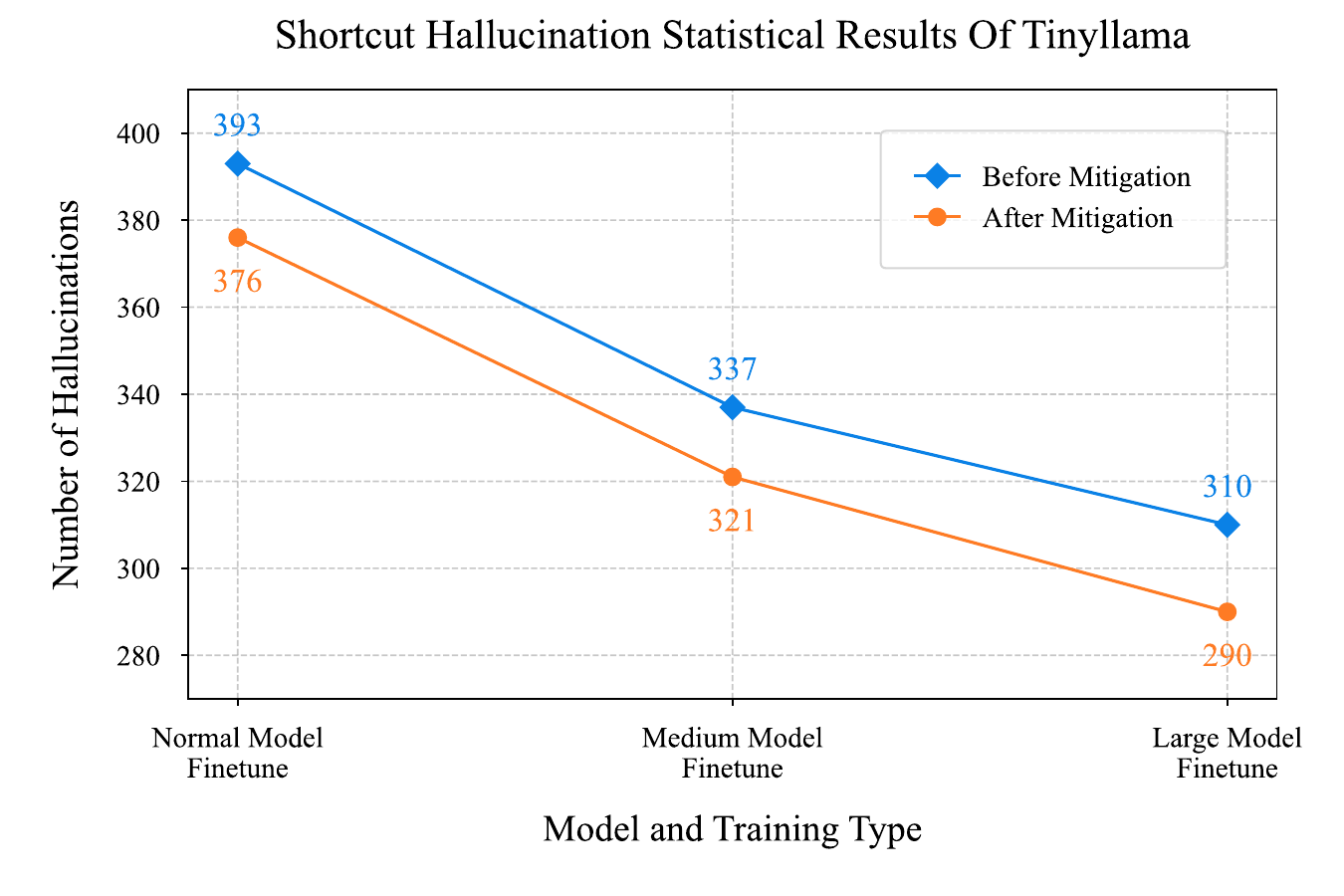} \hfill
  \includegraphics[width=0.48\linewidth]{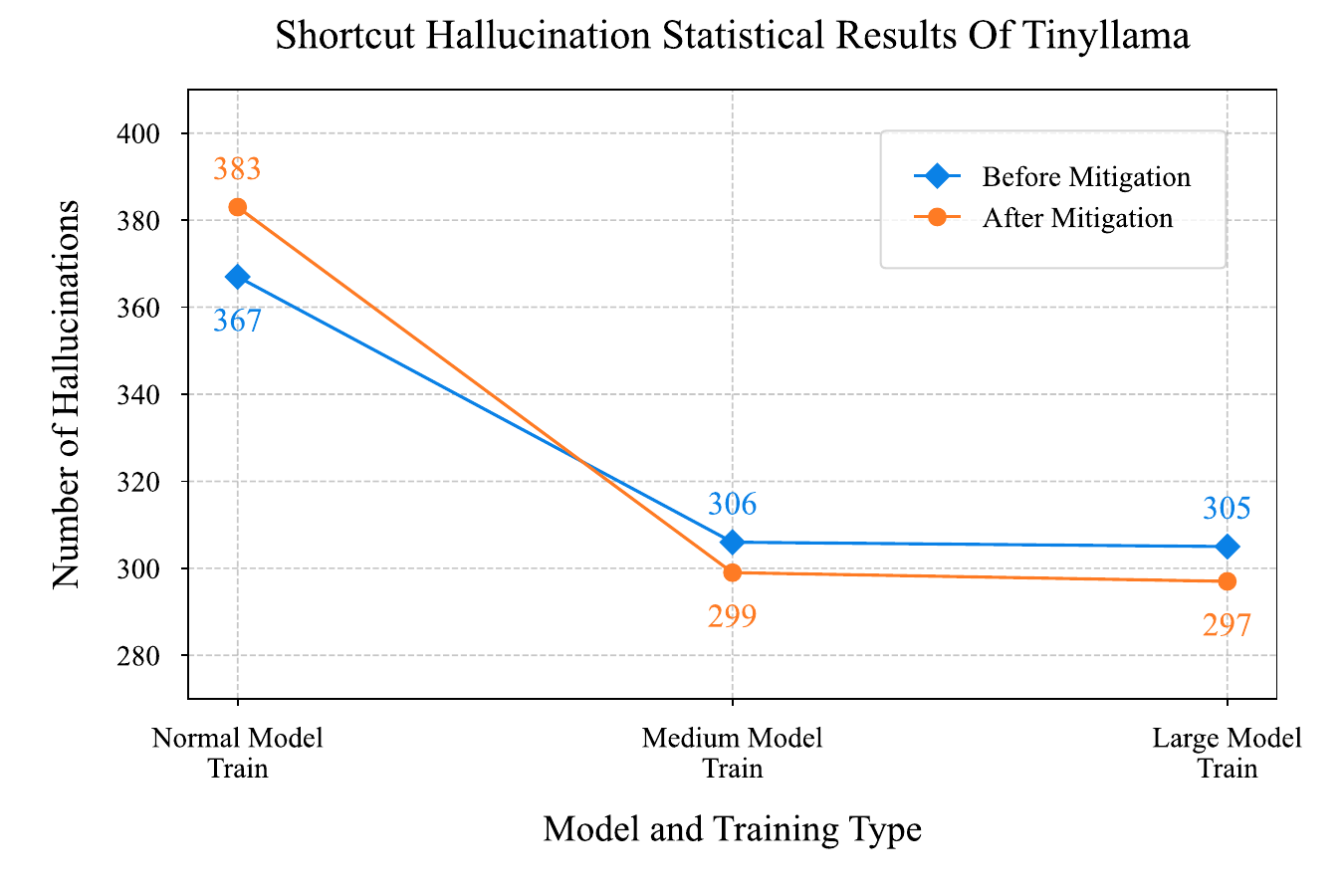}
  \caption {The number of Knowledge-Shortcut hallucination in CQA tasks before and after mitigation}
  \label{fig: detection number}
\end{figure*}
\begin{table*}[htbp!]
\centering
\begin{tabular}{llcccccc}
\toprule
\textbf{Type} & \textbf{Parameter} & \multicolumn{3}{c}{\textbf{After mitigation}} & \multicolumn{3}{c}{\textbf{Before mitigation}} \\
\cmidrule(lr){3-5} \cmidrule(lr){6-8}
& & \textbf{Jaccard} & \textbf{TF-IDF} & \textbf{Pre-train} & \textbf{Jaccard} & \textbf{TF-IDF} & \textbf{Pre-train} \\
\midrule
\multirow{6}{*}{\textbf{fine-tune}} 
& large   & \textbf{1649} & \textbf{1656} & 2400 & 1646 & 1652 & 2400 \\
&                & \textbf{0.62541} & \textbf{0.61355} & \textbf{0.79622} & 0.62188 & 0.61136 & 0.76381 \\
& medium  & \textbf{1388} & \textbf{1399} & 2400 & 1368 & 1380 & 2400 \\
&                & \textbf{0.51048} & \textbf{0.50020} & \textbf{0.69581} & 0.50859 & 0.49707 & 0.69392 \\
& normal  & \textbf{956} & \textbf{979} & 2400 & 952 & 975 & 2400 \\
&                & \textbf{0.32638} & \textbf{0.32113} & \textbf{0.56459} & 0.32576 & 0.31948 & 0.56337 \\
\midrule
\multirow{6}{*}{\textbf{train}} 
& large   & \textbf{1852} & \textbf{1861} & 2400 & 1838 & 1849 & 2400 \\
&                & \textbf{0.71533} & \textbf{0.70426} & \textbf{0.83314} & 0.71288 & 0.69956 & 0.82451 \\
& medium  & \textbf{1737} & \textbf{1745} & 2400 & 1699 & 1709 & 2400 \\
&                & \textbf{0.66219} & \textbf{0.65106} & \textbf{0.79507} & 0.64394 & 0.63185 & 0.78482 \\
& normal  & 1341 & 1352 & 2400 & \textbf{1373} & \textbf{1385} & 2400 \\
&                & 0.49678 & 0.48617 & 0.68270 & \textbf{0.50845} & \textbf{0.49828} & \textbf{0.68748} \\
\bottomrule
\end{tabular}
\caption{Results with related test, mitigation and before mitigation for various nanoGPT parameters. The integer above each row represents the number of non-zero and non-empty result rows (out of a total of 2400), while the decimal below indicates the overall average similarity. Values in \textbf{bold} denote significant results.}
\label{tab:after mitigation evaluate}
\end{table*}
\subsection{Experiment 2: Mitigation--Effectiveness Before and After Applying the High Similarity Pruning Algorithm}

Using the same training methods and parameters, we conducted CQA tasks on the models before and after applying the mitigation strategy.  We first evaluate whether the 0.4\% reduction in training data induced by the mitigation strategy would adversely affect model training performance. While the primary objective of this mitigation strategy is to reduce knowledge shortcut-induced model hallucinations, we aim to ensure that its implementation does not compromise model performance on CQA tasks. We employ two coarse-grained evaluation metrics: the count of non-zero non-empty similarity rows and average similarity score. The strategy is considered effective if these metrics demonstrate comparable or slightly improved performance post-implementation.

As shown in Table \ref{tab:after mitigation evaluate}, across nearly all training configurations and model scales of nanoGPT, the mitigation strategy generally yields marginal improvements in both metrics. An exceptional performance decline observed in the GPT-2 (124M) model under one specific training configuration will be analyzed in the section \ref{limitations}.

The reduction in training data directly leads to a decrease in training time, improving the efficiency of generative model training or fine-tuning. The reduced data did not impact testing on the \emph{related} test set. In the other experiments presented in Table \ref{tab:unrelated test after mitigation evaluate}, we also explored the impact on the \emph{unrelated} test set, where the fluctuations in both metrics remained minimal within the proportion of data reduction (For the unrelated test set, the train method is not meaningful, so we focus only on the fine-tune method). Furthermore, after applying the mitigation strategy, models trained on mixed datasets showed a decreasing trend in the proportion of "less" labels when compared to before mitigation in appendix \ref{after mitigation result}. 

\begin{table*}[htbp!]
\centering
\begin{tabular}{llcccccc}
\toprule
\textbf{Type} & \textbf{Parameter} & \multicolumn{3}{c}{\textbf{After mitigation}} & \multicolumn{3}{c}{\textbf{Before mitigation}} \\
\cmidrule(lr){3-5} \cmidrule(lr){6-8}
& & \textbf{Jaccard} & \textbf{TF-IDF} & \textbf{Pre-train} & \textbf{Jaccard} & \textbf{TF-IDF} & \textbf{Pre-train} \\
\midrule
\multirow{6}{*}{\textbf{fine-tune}} 
& large   & 1090 & 1126 & 2056 & \textbf{1118} & \textbf{1160} & 2056 \\ 
&                & 0.41782 & 0.37675 & \textbf{0.64697} & \textbf{0.42160} & \textbf{0.37675} & 0.64240 \\ 
& medium  & \textbf{976}  & \textbf{1023} & 2056 & 956  & 1006 & 2056 \\ 
&                & 0.36990 & \textbf{0.33733} & 0.60472 & \textbf{0.37286} & 0.33688 & \textbf{0.60864} \\ 
& normal  & \textbf{632}  & \textbf{675}  & 2056 & 601  & 648  & 2056 \\ 
&                & \textbf{0.22673} & \textbf{0.20910} & \textbf{0.49422} & 0.21640 & 0.19913 & 0.48892 \\ 
\midrule
\end{tabular}
\caption{Results with unrelated test, before and after mitigation. The integer above each row represents the number of non-zero and non-empty result rows (out of a total of 2056), while the decimal below indicates the overall average similarity. Values in \textbf{bold} denote significant results.}
\label{tab:unrelated test after mitigation evaluate}
\end{table*}
\subsection{Experiment 3: Detection--Reduction of Knowledge-Shortcut hallucinations}
To evaluate the effectiveness of the mitigation strategy from a finer-grained perspective, we employed the knowledge-shortcut hallucination fusion detection method. Specifically, we directly counted the number of knowledge-shortcut hallucinations in the test set generated by models of different parameter scales and training methods before and after applying the mitigation strategy. This straightforward approach provides a clear demonstration of the mitigation strategy's effectiveness.

Our detections show that in large-parameter fine-tuning models, the mitigation strategy performs exceptionally well in suppressing knowledge-shortcut hallucinations. Even for the training method, the strategy proved effective in reducing hallucinations, validating the method's efficacy in mitigating knowledge-shortcut hallucinations in generative models. The results are presented in Figure \ref{fig: detection number}.

We provide a reproducible set of repeated experimental results in Appendix \ref{reproducible repeated experiments}. The training and generation code has been open-sourced on GitHub. The results from this set of experiments align well with those presented in the main text, indirectly validating the robustness of our method.


\section{Related Works}
Significant progress has been made in the study of hallucinations in LLMs. Xu\cite{xu2024hallucination} argue that eliminating hallucinations in LLMs is impossible. Numerous techniques have emerged to mitigate LLM hallucinations, with notable approaches including Retrieval-Augmented Generation (RAG)\cite{lewis2020retrieval}, knowledge retrieval\cite{varshney2023stitch}, CoNLI\cite{lei2023chain}, and CoVe\cite{dhuliawala2023chain}. Our aim is to clarify the underlying causes of knowledge-shortcut hallucinations and minimize such hallucinations as much as possible.

Shortcut learning is a critical area of research in LLM hallucinations, with knowledge shortcuts representing the manifestation of shortcut learning at the data level. Geirhos\cite{geirhos2020shortcut} and Du\cite{du2023shortcut} suggest that dataset bias is the starting point of shortcut learning, and many excellent works have focused on alleviating shortcut learning from the data perspective, such as identifying biased sentencesLei\cite{lei2022sentence}, data shortcuts\cite{friedman2022finding}, or replacing datasets with more balanced ones\cite{tang2023large}. Tang's research\cite{tang2021mitigating} indicates that fine-tuned language models can learn and even amplify biases present in the training datasets, leading to poor performance in downstream tasks, which aligns with our experimental findings. Despite these advancements, existing works have yet to fully explore all the ways in which dataset bias can manifest. Our goal is to conduct an in-depth study of one type of hallucination triggered by correct, defect-free data sources in large models and propose a feasible method for mitigating and predicting such illusions.

\section{Conclusion}
In conclusion, we have conducted a finer-grained study of a specific type of hallucination originating from the data perspective and proposed a novel method for mitigating this hallucination, along with a fusion detection method for such hallucinations. Our approach demonstrates through experiments that, when handling specific question-answering tasks, it can significantly reduce knowledge-shortcut hallucinations in the fine-tuning process while maintaining the performance of generative models and stabilizing answer similarity. This provides a new paradigm for addressing hallucinations in generative models.

\section{Limitations}
\label{limitations}
\textbf{Normal parameter scale results}. In the experiment shown in Figure \ref{fig: detection number}, the mitigation effect on the nanoGPT (124M) model was relatively poor. This phenomenon persisted in repeated experiments (Figure \ref{fig: new number of ksh}), suggesting a potential explanation. Given the small parameter scale and limited training data, the model may struggle to learn the patterns of knowledge shortcuts effectively. As a result, applying the mitigation strategy does not yield significant improvements. This observation indicates that our mitigation approach is better suited for larger-scale models, aligning with the experimental results observed in large parameter models.

\textbf{Runtime and applicable tasks}. Our mitigation strategy demonstrated outstanding performance in fine-tuning tasks, with a stable and significant reduction in knowledge-shortcut hallucinations. This suggests that the strategy is more suitable for fine-tuning rather than pretraining tasks. From a data scale perspective, pretraining datasets are typically vast, whereas fine-tuning datasets are relatively smaller. As a result, the computational overhead introduced by our mitigation strategy is entirely acceptable in fine-tuning scenarios.

\textbf{Detection method applicability}. Unlike the general mitigation strategies at the data preprocessing level, our knowledge shortcut hallucination detection method is specifically designed for CQA tasks and is data-dependent. As such, we did not assess the superiority of this method; rather, it serves as an evaluation technique for the number of knowledge shortcut hallucinations before and after applying our mitigation strategy. Due to its data dependency, this detection method is also applicable to fine-tuning tasks.

\textbf{Chain-of-Thought technology}. We have also tested floating-point comparison issues on the latest commercial large models utilizing "chain-of-thought" (CoT) reasoning, such as ChatGPT o1 and DeepSeek R1. Although they ultimately provided the correct answers, doubts arose during the reasoning process. ChatGPT o1, for instance, initially gave an incorrect answer but corrected itself in the subsequent reasoning steps. Therefore, CoT technology represents a potential approach for mitigating all factual hallucinations. However, the phenomenon of knowledge shortcuts may not directly affect the final outcome, yet it still misleads the model's reasoning process.
\section{Acknowledgements}

\bibliography{custom}

\appendix

\section{Example}
\label{example}
In Table \ref{tab:example before and after mitigation}, the word marked in red in \emph{Answer3} is identified because it repeatedly appears in the high similarity group between the test input (CQ) and other datasets. This is further detailed in Table \ref{tab:example for lone context}. The first three rows of the table indicate that, in the top 50 rows with the highest Jaccard similarity to this CQ from the \emph{trivia(data3)} dataset, words \textbf{\emph{\{gland\}}}, \textbf{\emph{\{adrenal, gland\}}}, and \textbf{\emph{\{gland\}}} were found in rows $17$, $18$, and $49$ (with indices $1326$, $11791$, and $1303$ in the data3 dataset, respectively). Therefore, we define the incorrect answer containing the red-marked word as a knowledge-shortcut hallucination.

\begin{table}
  \centering
  \begin{tabular}{lc}
    \hline
    \textbf{Model Name} & \textbf{Answer} \\
    \hline
    \verb|Chatgpt-4|           & {$\times$}           \\
    \verb|Chatgpt-4o|          & {$\times$}           \\
    \verb|Gemini Advanced|     & {$\times$}           \\
    \verb|Claude|              & {$\times$}           \\
    \verb|Kimi|                & {$\times$}            \\
    \verb|Cici|                & {$\times$}           \\
    \verb|ERNIE Bot|           & {\checkmark}           \\
\hline
  \end{tabular}
  \caption{Large Model Floating Point Comparison}
  \label{tab:counts}
\end{table}

\section{Methodology}
\subsection{Metrics of Text Similarity}
\label{metrics of similarity}
This paper aims to develop a framework for detecting and mitigating knowledge-shortcut hallucinations and to validate the effectiveness of the proposed method. From the perspective of data, the root cause of knowledge-shortcut hallucinations is intuitively reflected in the presence of high textual similarity within the training data. To quantify textual similarity, we employ three mathematical approaches:

\textbf{Jaccard similarity}. Jaccard similarity is the most straightforward measure of the overlap between two sets. It calculates the ratio of the intersection to the union of the sets, providing a naive yet effective way to assess similarity. The numerator denotes the intersection of sets A and B, and the denominator denotes the union of sets A and B.
\begin{equation}
  \label{Jaccard_sim}
  Jaccard_{sim}=\frac{|A\bigcap B|}{|A\bigcup B|}
\end{equation}
\textbf{TF-IDF similarity}. TF-IDF(Term Frequency-Inverse Document Frequency) similarity leverages statistical measures of word importance across documents, enabling a more nuanced comparison that considers term frequency and discriminative power. The $t$ in the formula denotes the word and $d$ denotes the document. So we can get $TF-IDF_{sim}$
\begin{equation}
  \label{tfidf_sim}
  TF-IDF_{sim}(t,d)=TF(t,d)\cdot IDF(t)
\end{equation}
\begin{equation}
  \label{tf}
  TF(t,d)=\frac{count(t,d)}{\sum_{k\in d}count(k,d)}
\end{equation}
\begin{equation}
  \label{idf}
  IDF(t)=log\frac{N}{1+DF(t)}
\end{equation}
\textbf{Pre-trained model-based similarity}. A Pre-trained model-based similarity measure uses pre-trained language models to compute semantic similarity between text pairs. This method captures contextual and latent relationships in text, providing a more sophisticated and accurate measure compared to traditional approaches. Finally, we choose the paraphrase-miniLM-v12-v2\footnote{\url{https://huggingface.co/sentence-transformers/paraphrase-MiniLM-L12-v2}} of the sentence-transformers library, because it performs fast, accurate sentence similarity evaluation.

We utilize the above three metrics to measure the similarity between the generated responses and the correct answers, leveraging their respective advantages. Jaccard similarity provides a simple and intuitive method for quickly assessing the magnitude of similarity between two texts. TF-IDF similarity incorporates the influence of term frequency, reducing the impact of high-frequency words on sentence similarity. Pre-trained model-based similarity evaluates the similarity from a semantic perspective, offering fine-grained corrections for discrepancies, such as those between \emph{"4"} and \emph{"four."} By combining these three metrics, we achieve a multidimensional evaluation of sentence similarity. 

We also performed several engineering optimizations in the code. Given that the generated responses in the CQA task are relatively short, variations in singular and plural forms of nouns, verb conjugations, and adjective-adverb transformations could significantly impact Jaccard similarity and TF-IDF similarity. To address this, we utilized the \emph{nltk\footnote{\url{https://www.nltk.org/}}} library to implement a lemmatization method, improving the accuracy of Jaccard and TF-IDF similarity measurements. This refinement enhances the granularity of overall text similarity evaluation metrics.

\subsection{The Defination of High Similarity Group}
\label{HS group}
\begin{figure}[t]
  \includegraphics[width=1.1\linewidth]{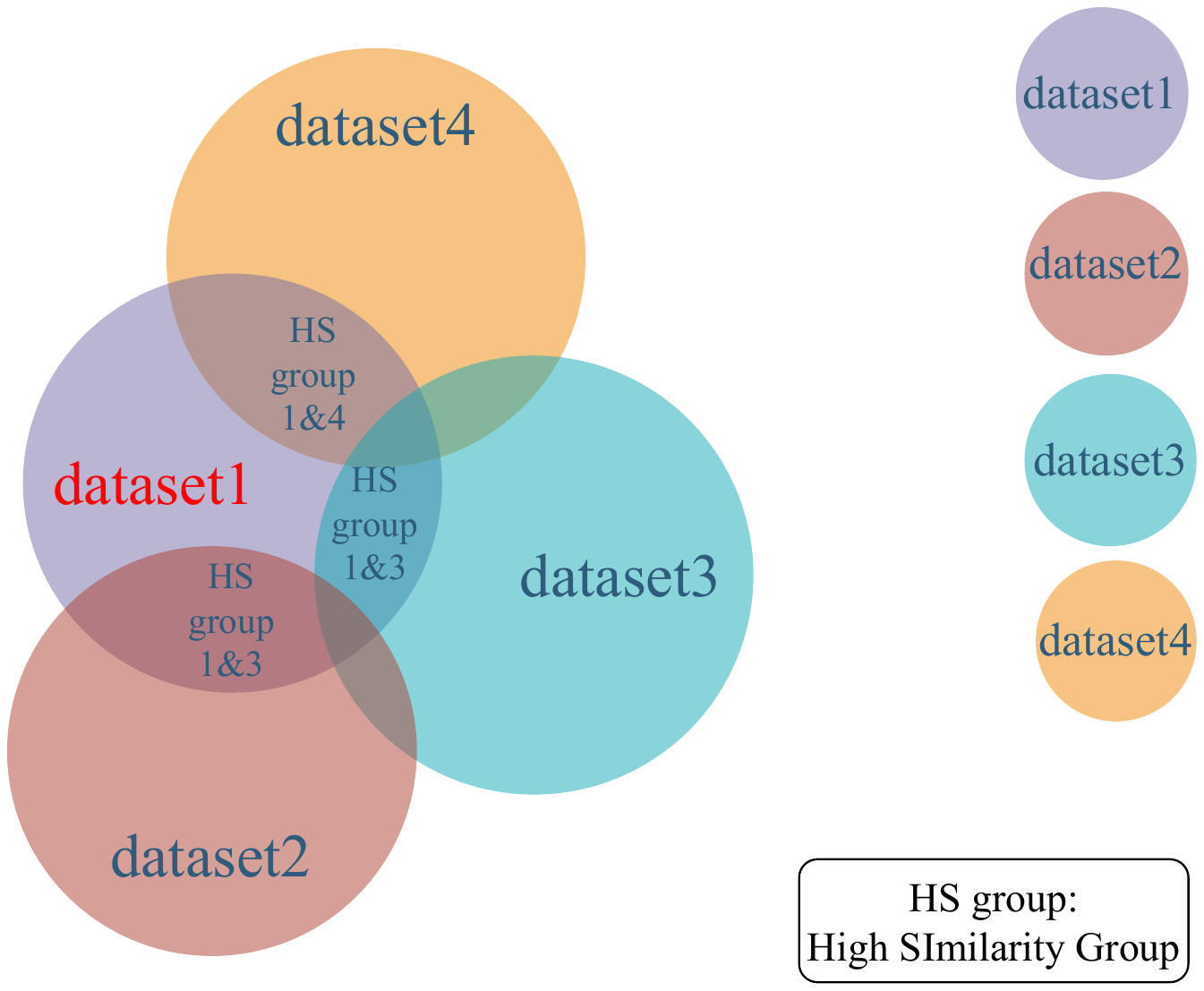}
  \caption{An example of data high similarity}
  \label{fig:overview}
\end{figure}

Datasets used in pretraining or fine-tuning often consist of multiple semantically distinct sources, such as mathematics, art, and agriculture. Our focus is on subsets of high textual similarity within these sources. As shown in Figure \ref{fig:overview}, for example, high similarity between \emph{data1} and subsets of \emph{data2}, \emph{data3}, and \emph{data4} forms what we call \emph{High Similarity Group} (HS group), represented as \emph{HS group 1\&2}, \emph{HS group 1\&3}, and \emph{HS group 1\&4}. These high similarity texts can convey different meanings, misleading generative models during training or fine-tuning. This increases the risk of knowledge-shortcut hallucinations when queries are input. 

\subsection{Pseudocode of Mitigation and Detection}
\label{sec:appendix pseudocode}

\begin{algorithm}[ht]  
  \caption{Mitigation: High Similarity Pruning Algorithm }  
  \label{alg:Framwork}  
  \begin{algorithmic}[1]
    \Require  
        $(data_1,...,data_n)$;
        $K_1,K_2,\alpha_1,\alpha_2$
    \Ensure 
        Mitigation strategy
    
    \For{$k=1$ to $n$}
        \For{$i=0$ to row($data_k$)}
            \For{$j\neq k$ to $n$}
            
                $T_{i,j}(CQA_{G_{HF},G_{HV}})\leftarrow K_1,data_j$;
                
                $G_{j,HF},G_{j,HV}\leftarrow T_{i,j}(CQA_{G_{HF},G_{HV}})$;

            \EndFor
            
        $R_{k,j}\leftarrow Set(\alpha_1 G_{j,HF}+\alpha_2 G_{j,HV})$;
        \EndFor
        
        $R_{all}\leftarrow Set(R_{k,j}),K_2$;
    \EndFor
    
  $(data_1,...data_n)'\leftarrow(data_1,data_n)-R_{all}$;
           
  \end{algorithmic}  
\end{algorithm}

\begin{algorithm}[ht]  
  \caption{Knowledge Shortcut Hallucination Detection}  
  \label{alg:Framework}  
  \begin{algorithmic}[1]  
    \Require  
        Context-question pair $(CQ)$;   
        Similarity threshold $\alpha_3$;  
        $m=5$  
    \Ensure  
        Detection of knowledge-shortcut hallucination  
    \For{$j=0$ to $len(CQ)$}
    
        \State $T(CQA_{ij})\leftarrow (data_1,...data_n)$;
        \State $G_{HF},G_{HV}\leftarrow CQA_{ij}$;
        \For{$l = 1$ to $m$}  
            \State Generate response $A_l$;  
            \If{$\frac{\sum_{l=1}^m 1-Sim(A_o,A_{l})}{m} > \alpha_3$}  
            
                $S_o = Set(A_o) - Set(CQ)$;
                \If{$S_o$ is not empty} 
                    \State $flag \gets Set(A_o)$
                    \Statex \hspace{2cm} $\cap Set(CQA_{G_{HF},G_{HV}})$;

                    \If{$flag\neq \emptyset$}  
                        \State Return True;
                    \Else{ Return False;}
                    \EndIf
                \Else{ Return False;}
                \EndIf
            \Else{ Return False;}  
            \EndIf  
        \EndFor  
    \EndFor
  \end{algorithmic}  
\end{algorithm}

\section{Results of the Experiment}
\label{results}
\subsection{Evaluation after applying mitigation strategies}
\label{after mitigation result}
\begin{figure*}[t]
  \includegraphics[width=0.3\linewidth]{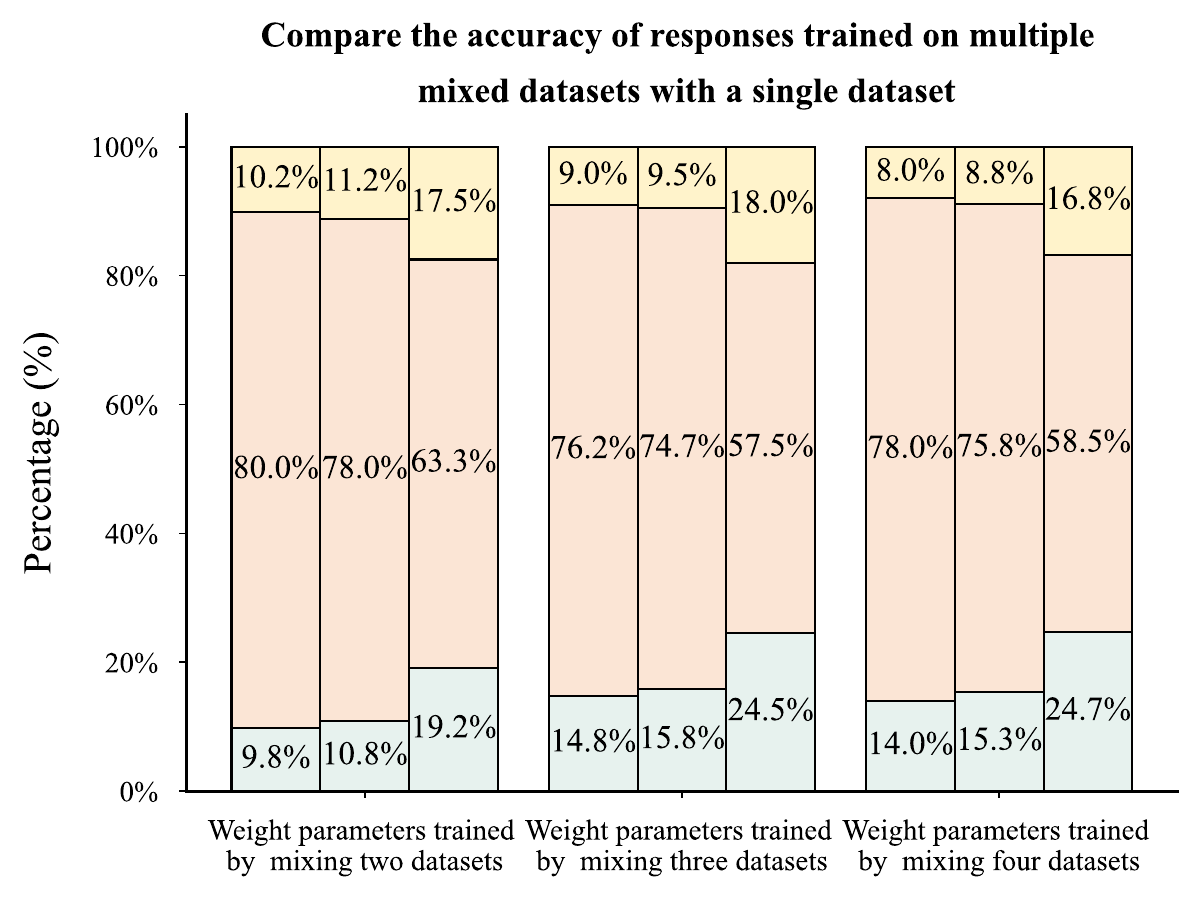} \hfill
  \includegraphics[width=0.3\linewidth]{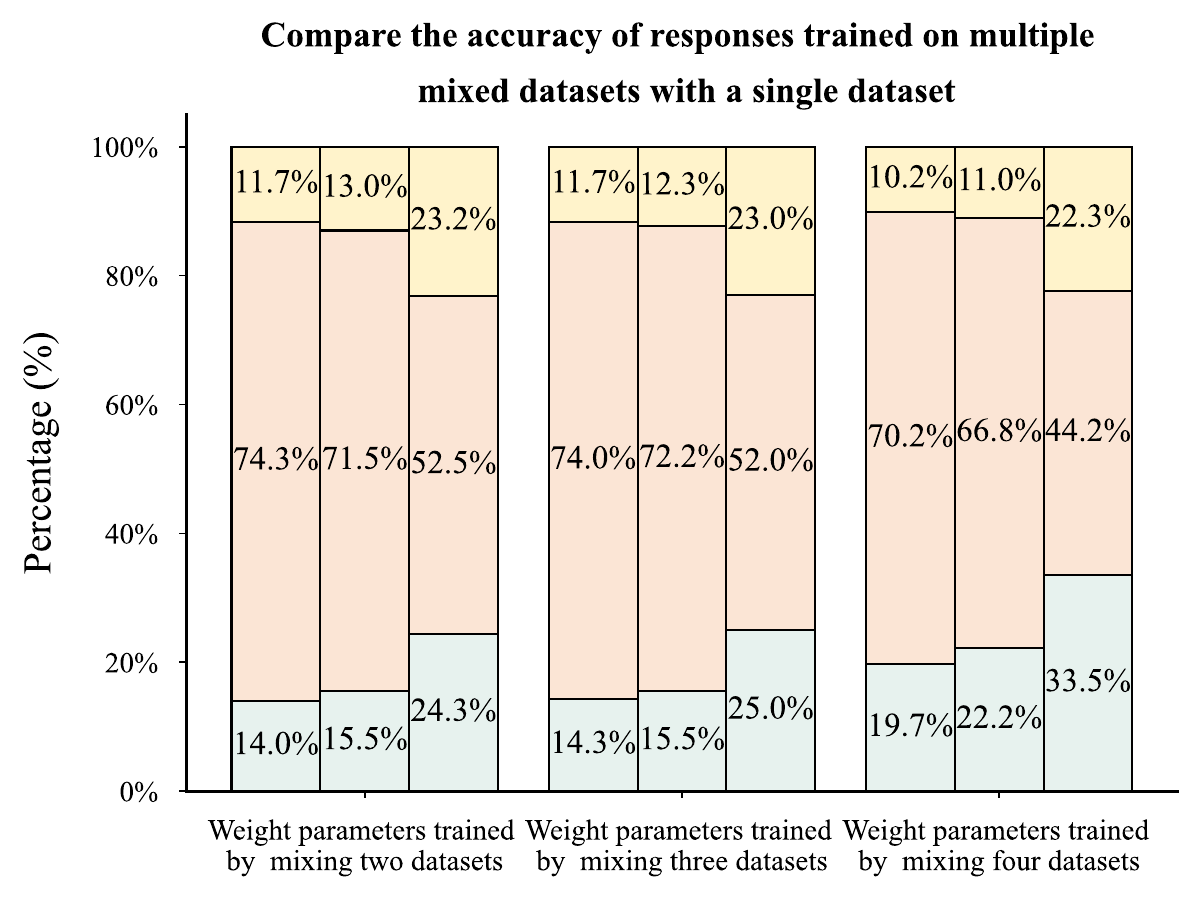} \hfill
  \includegraphics[width=0.36\linewidth]{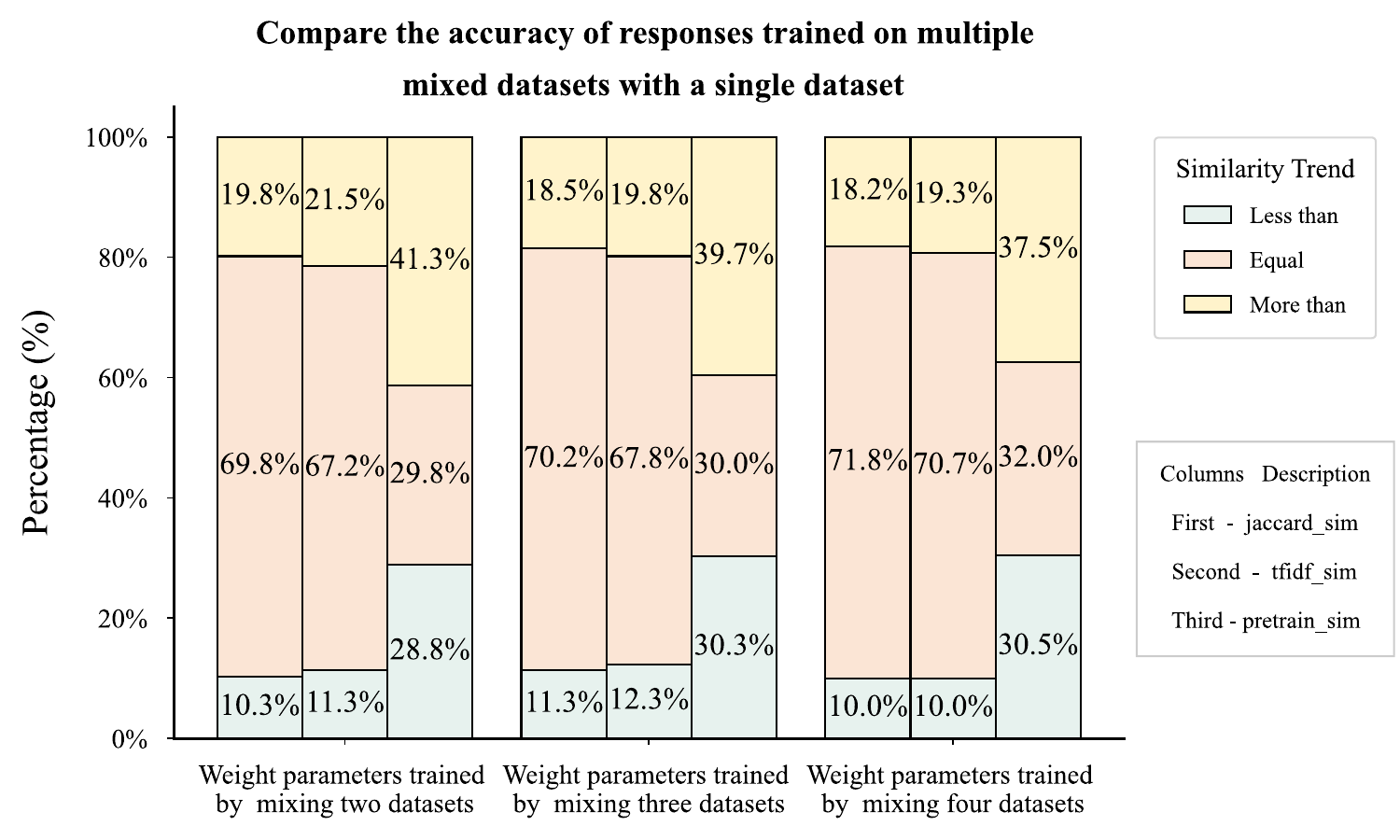}
  \caption {After mitigation, fine-tuning: nanoGPT large, medium, normal}
  \label{fig:ft-after}
\end{figure*}
\begin{figure*}[t]
  \includegraphics[width=0.3\linewidth]{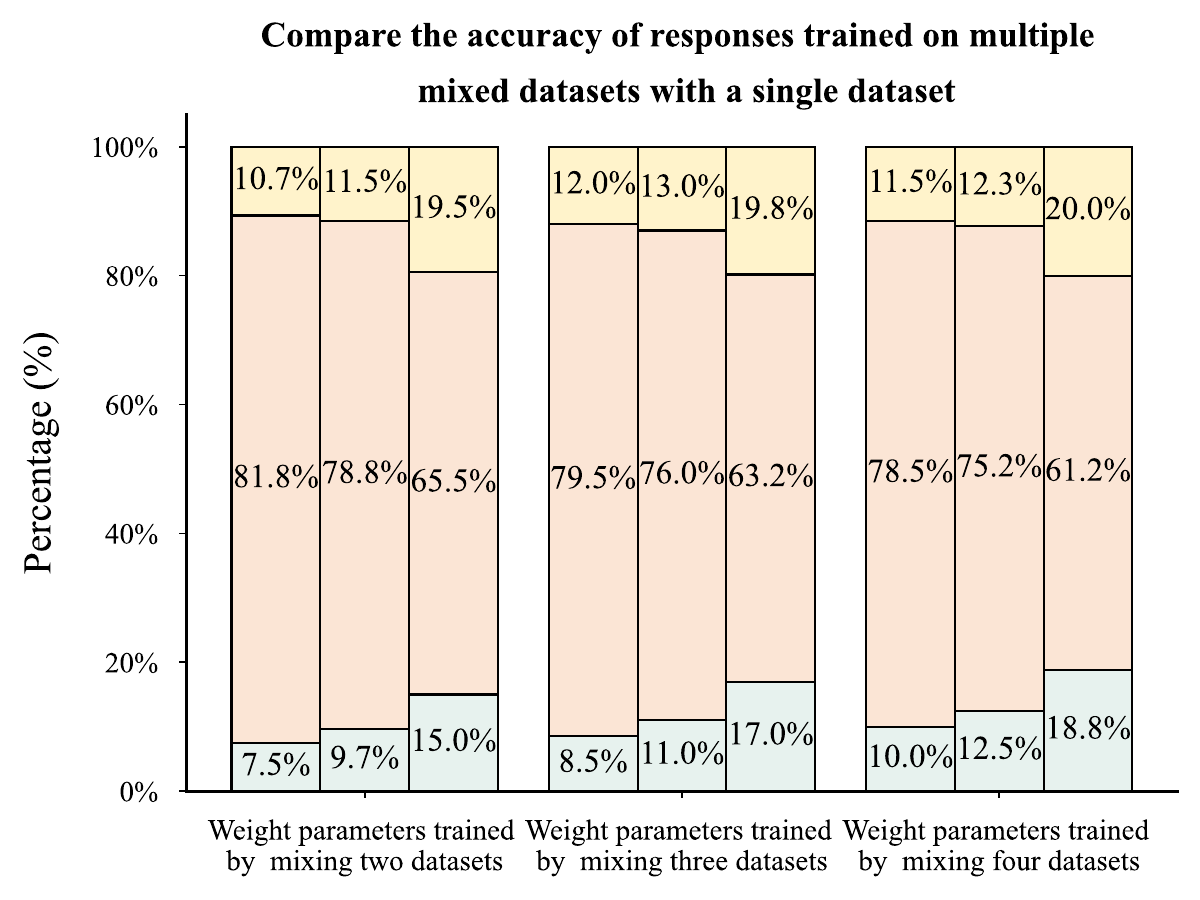} \hfill
  \includegraphics[width=0.3\linewidth]{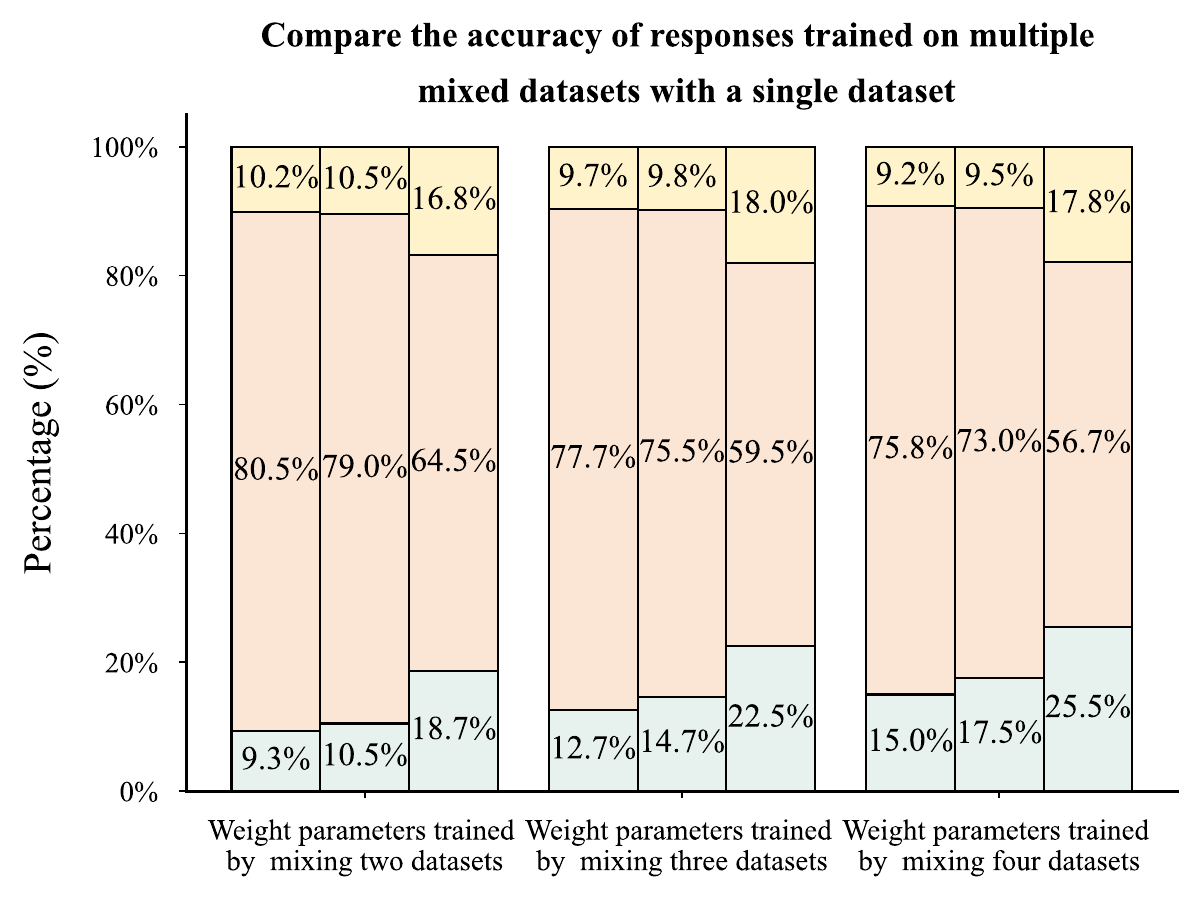} \hfill
  \includegraphics[width=0.36\linewidth]{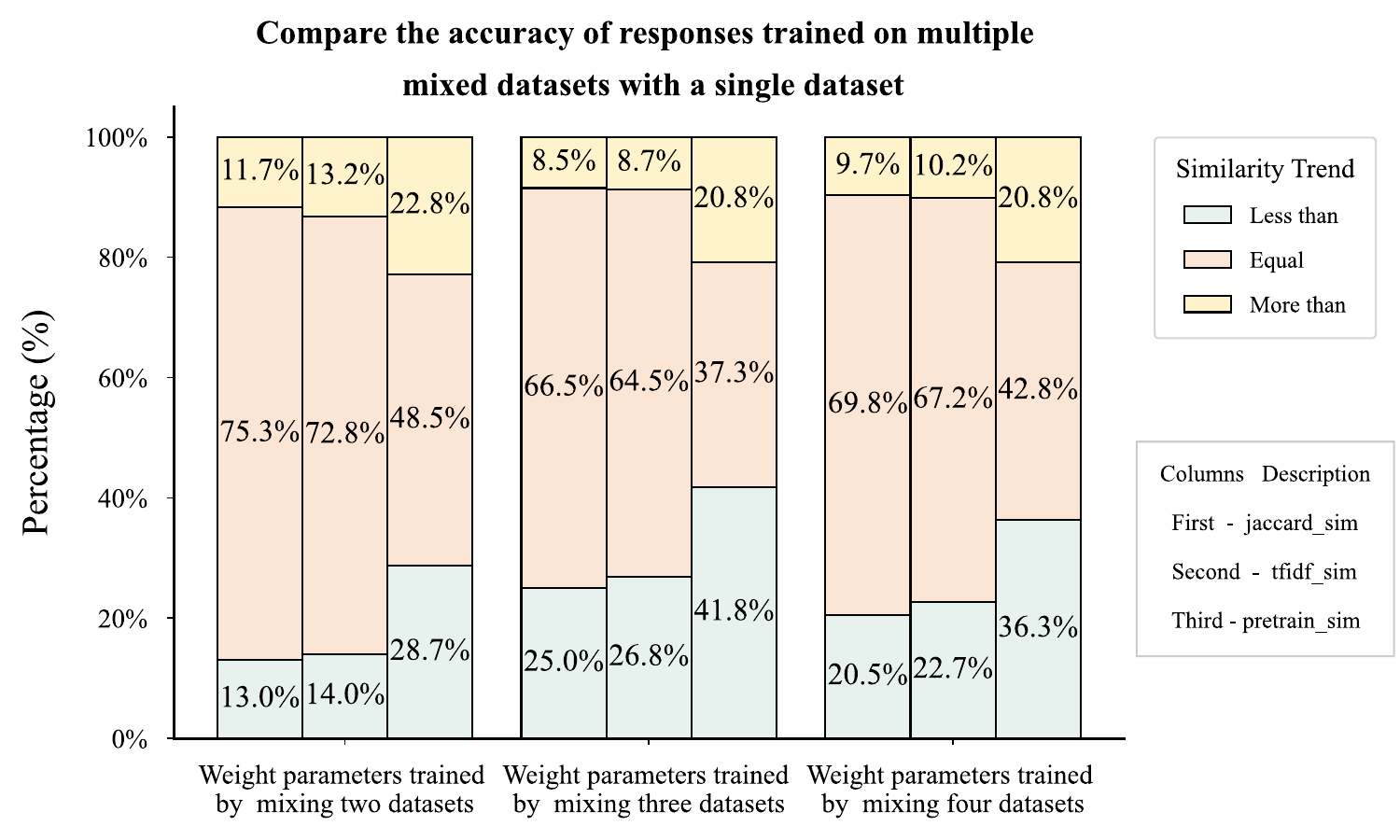}
  \caption {After mitigation, train : nanoGPT large, medium, normal}
  \label{fig:train-after}
\end{figure*}
Here, we show the evaluation results of different parameter-scale models and training methods on the same test set after applying the mitigation strategy. Figures \ref{fig:ft-after} and \ref{fig:train-after} show the evaluation results after applying the mitigation strategy, corresponding to Figures \ref{fig:ft-before} and \ref{fig:train-before}. A comparison reveals that the proportion of \emph{less} labels has generally decreased, while the proportion of \emph{more} labels has increased, leading to a more balanced and coordinated distribution. These findings indicate that our strategy effectively reduces the influence of unrelated datasets on the model's generated answers, thereby improving output quality.  

It is worth noting that the normal parameter scale of nanoGPT is only 124M, resulting in greater fluctuations in the experiments. Compared to larger parameter models, this introduces some instability, leading to deviations in certain results.

\subsection{Experiment Results of Tinyllama}
\label{tinyllama result}
To further investigate the generalization effectiveness of our proposed method, we conducted the same evaluation, mitigation, and prediction experiments using the TinyLlama model, employing only the fine-tuning approach. The results and trends remained consistent with those observed in nanoGPT, demonstrating the robustness and applicability of our method across different model architectures.  

\begin{figure}[t]
  \includegraphics[width=\linewidth]{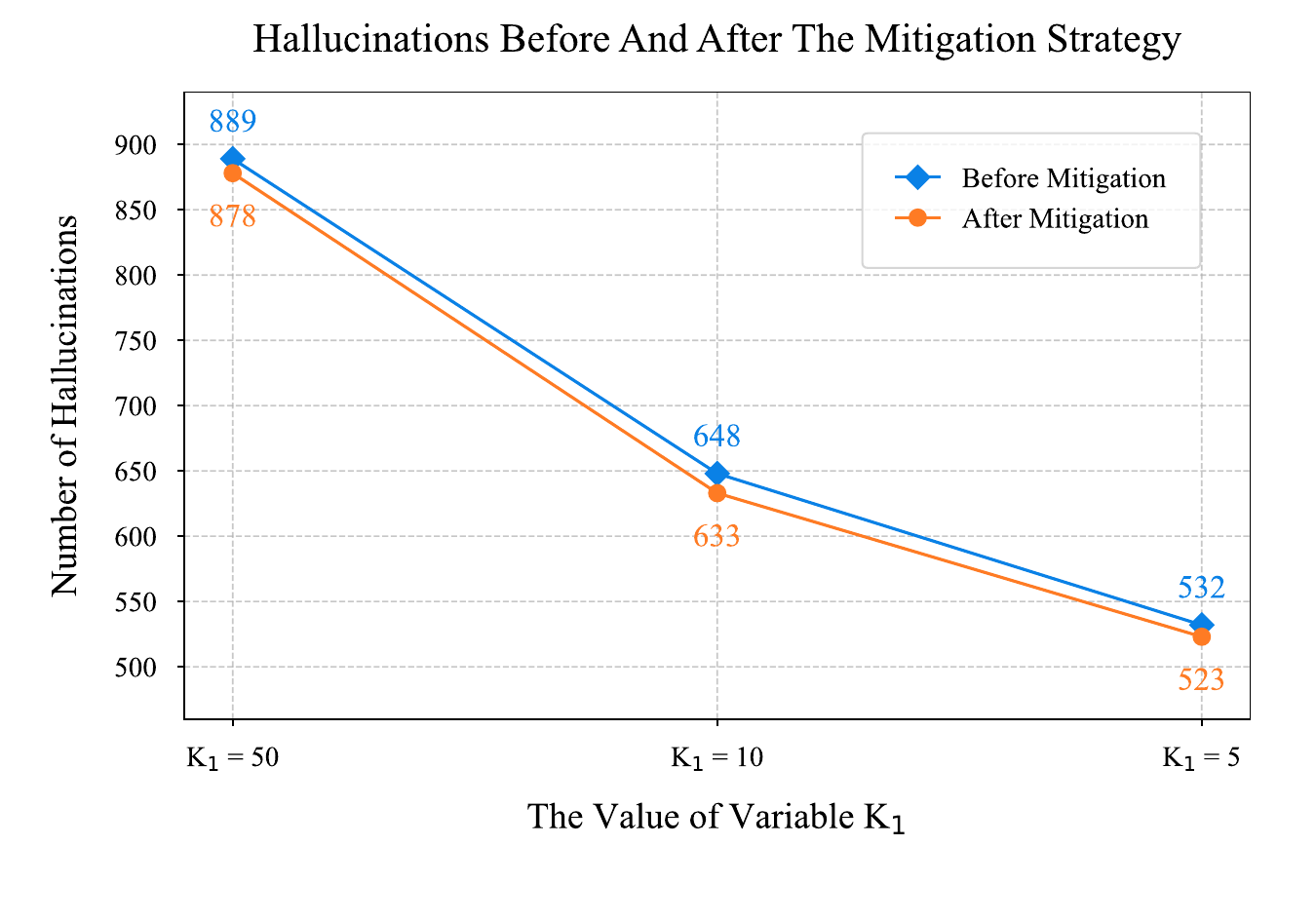}
  \caption {Effectiveness of Different $K_1$ Values on Tinyllama Knowledge-Shortcut Hallucination Detection}
  \label{fig:K_1 values}
\end{figure}

\begin{figure*}[htbp]
    \centering
    \begin{subfigure}[b]{0.48\textwidth} 
        \centering
        \includegraphics[width=\textwidth]{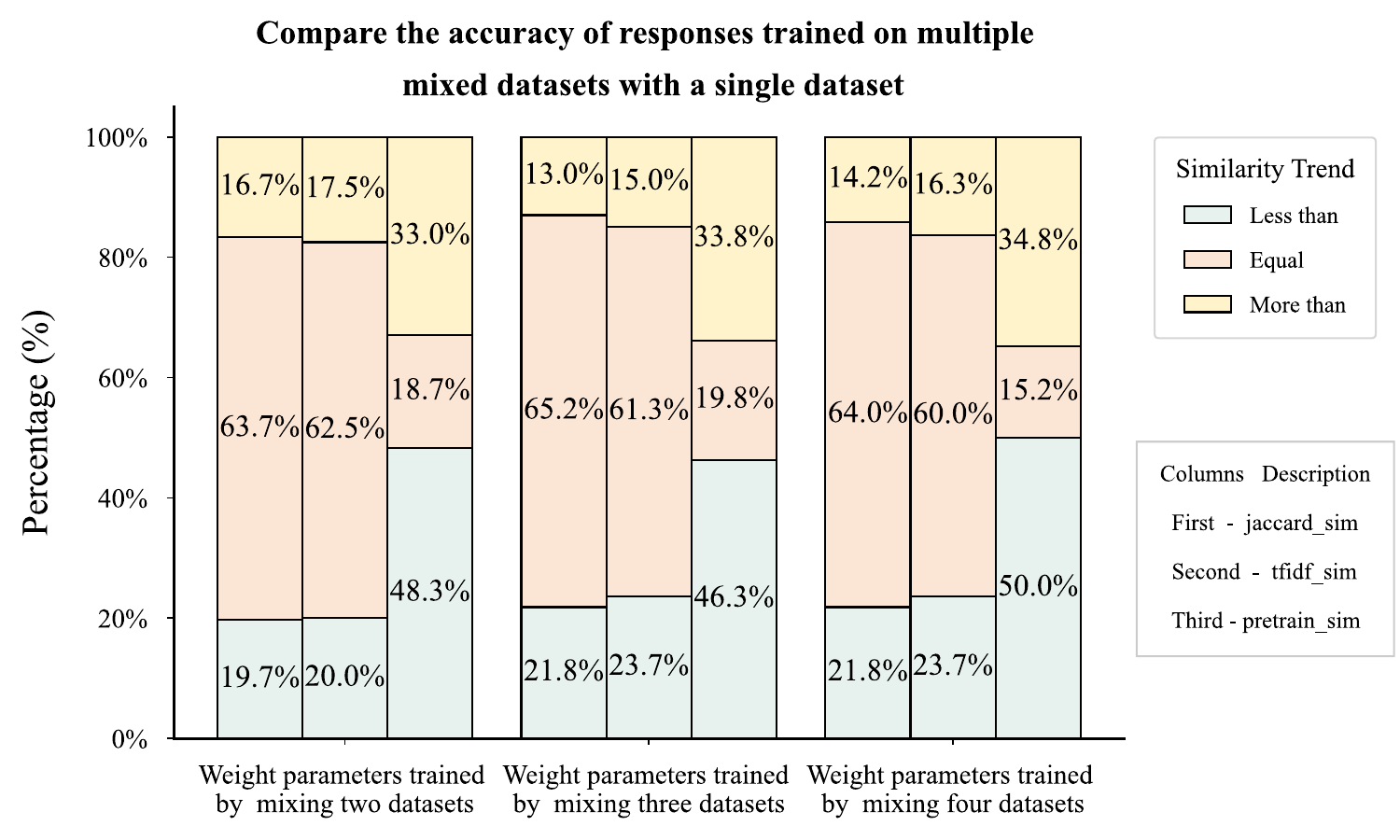} 
        \caption{Evaluation of Tinyllama, before mitigation} 
        \label{fig: tinyllama before_mitigation}
    \end{subfigure}
    \hfill
    \begin{subfigure}[b]{0.48\textwidth} 
        \centering
        \includegraphics[width=\textwidth]{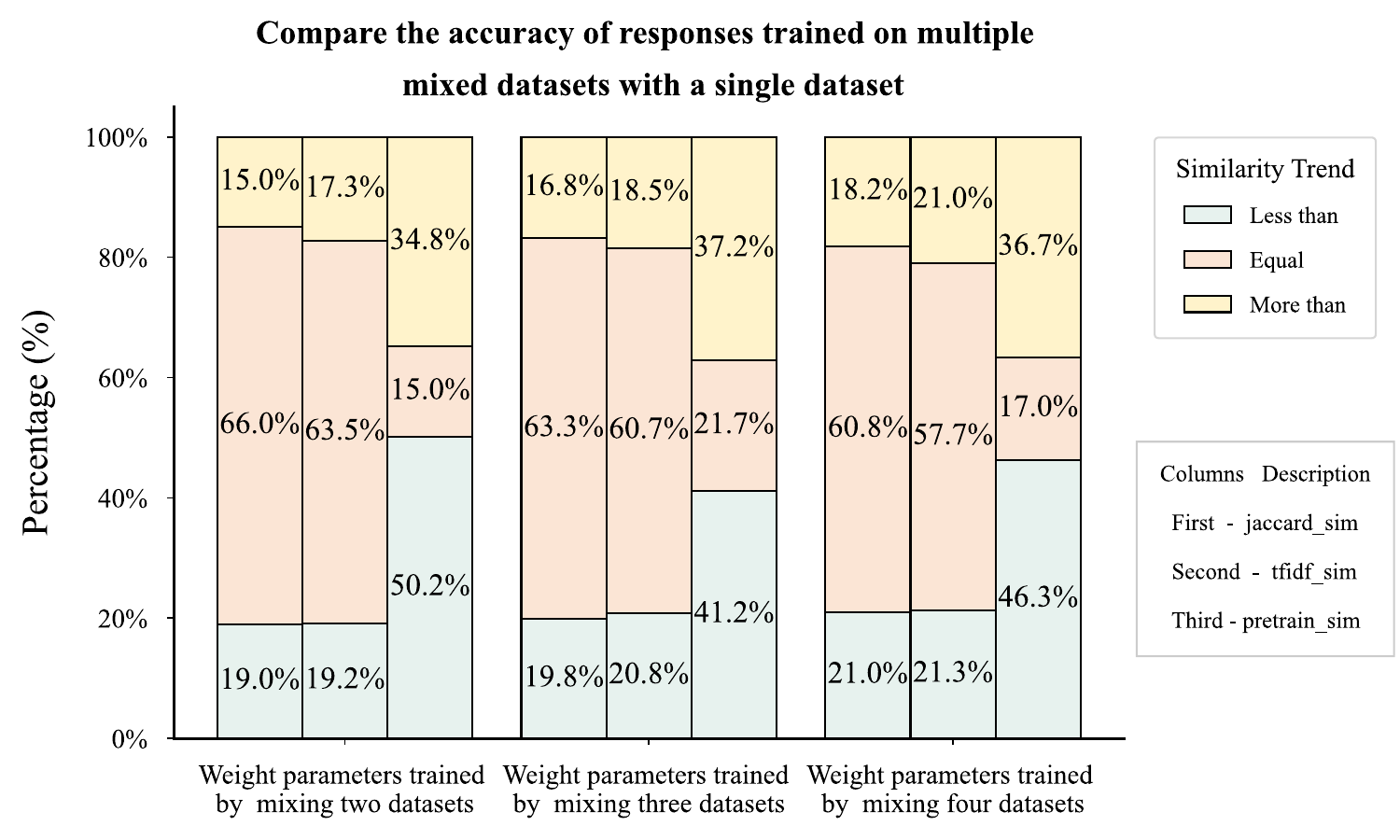} 
        \caption{Evaluation of Tinyllama, after mitigation} 
        \label{fig: tinyllama after_mitigation}
    \end{subfigure}

    \caption{(a) Before mitigation, (b) After mitigation. Illustration of the evaluation results with mitigation strategies.}
    \label{fig:tinyllama evaluation}
\end{figure*}

Figure \ref{fig:tinyllama evaluation} presents the evaluation results of TinyLlama before and after applying the mitigation strategy, while Table \ref{tab:tinyllama after mitigation sim} summarizes the macro-level performance metrics of TinyLlama’s generated responses under both conditions. Furthermore, in TinyLlama’s response generation process, we set the sampling parameter Top-k to 5. To ensure consistency, we maintained the hallucination detection parameter $K_1$ equal to \emph{Top-k} and conducted an ablation study to explore the impact of different $ K_1 $ values on the detection of knowledge-shortcut hallucinations. As shown in Figure \ref{fig:K_1 values}, after applying the mitigation strategy, the number of detected knowledge-shortcut hallucinations remained consistently lower than before across all $ K_1 $ values, with minimal variation in detection differences.


\begin{table*}[htbp!]
\centering
\begin{tabular}{llcccccc}
\toprule
\textbf{Type} & \textbf{Parameter} & \multicolumn{3}{c}{\textbf{After mitigation}} & \multicolumn{3}{c}{\textbf{Before mitigation}} \\
\cmidrule(lr){3-5} \cmidrule(lr){6-8}
& & \textbf{Jaccard} & \textbf{TF-IDF} & \textbf{Pre-train} & \textbf{Jaccard} & \textbf{TF-IDF} & \textbf{Pre-train} \\
\midrule
\multirow{1}{*}{\textbf{fine-tune}} 
& 1.1b   & \textbf{1273} & \textbf{1321} & 2400 & 1243 & 1280 & \textbf{2400} \\
&                & \textbf{0.39285} & \textbf{0.39105} & \textbf{0.68855} & 0.38596 & 0.38347 & 0.68374 \\
\midrule
\bottomrule
\end{tabular}
\caption{Results with mitigation and before mitigation for Tinyllama. Values in \textbf{bold} denote significant results.}
\label{tab:tinyllama after mitigation sim}
\end{table*}

\begin{figure*}[t]
  \includegraphics[width=0.48\linewidth]{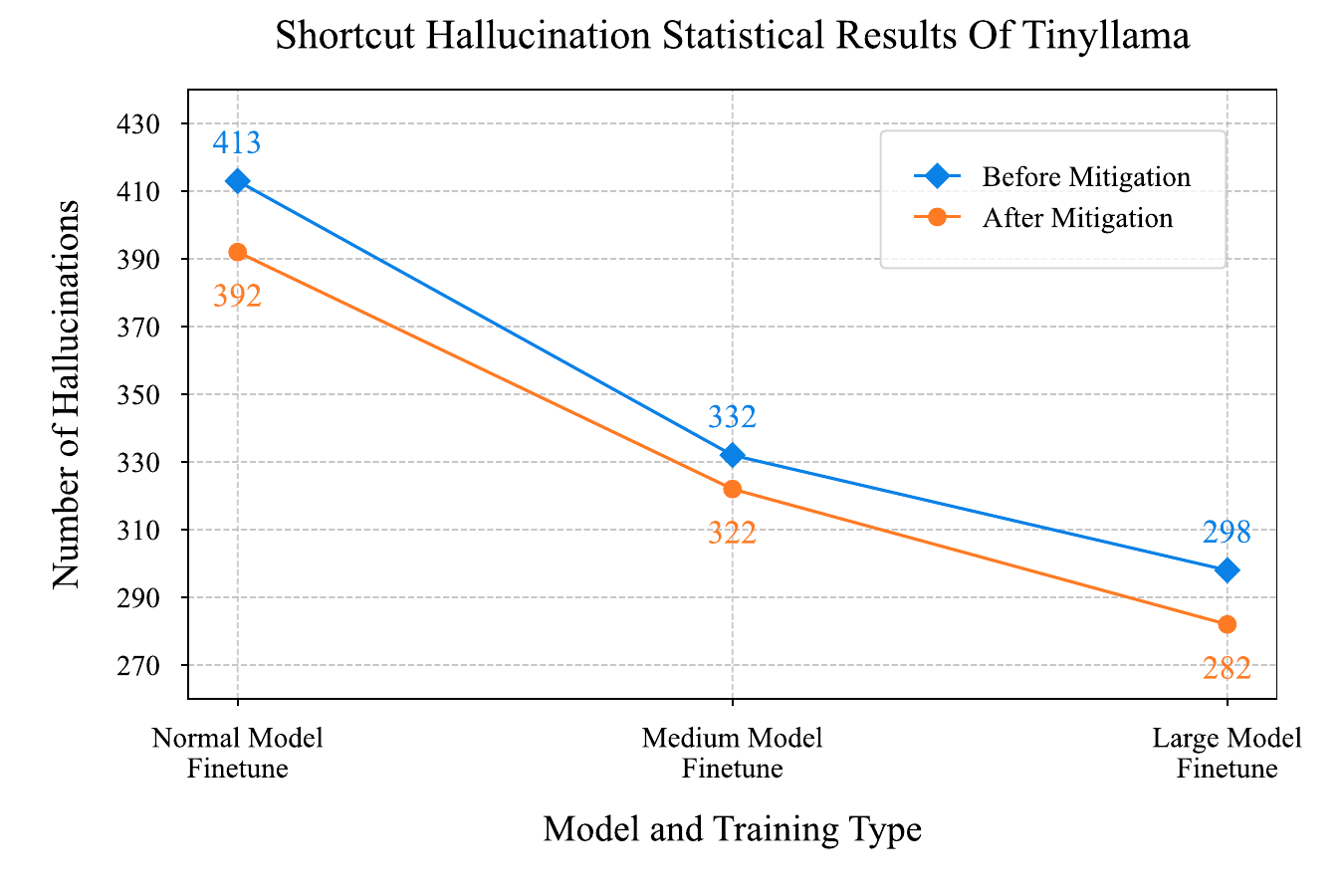} \hfill
  \includegraphics[width=0.48\linewidth]{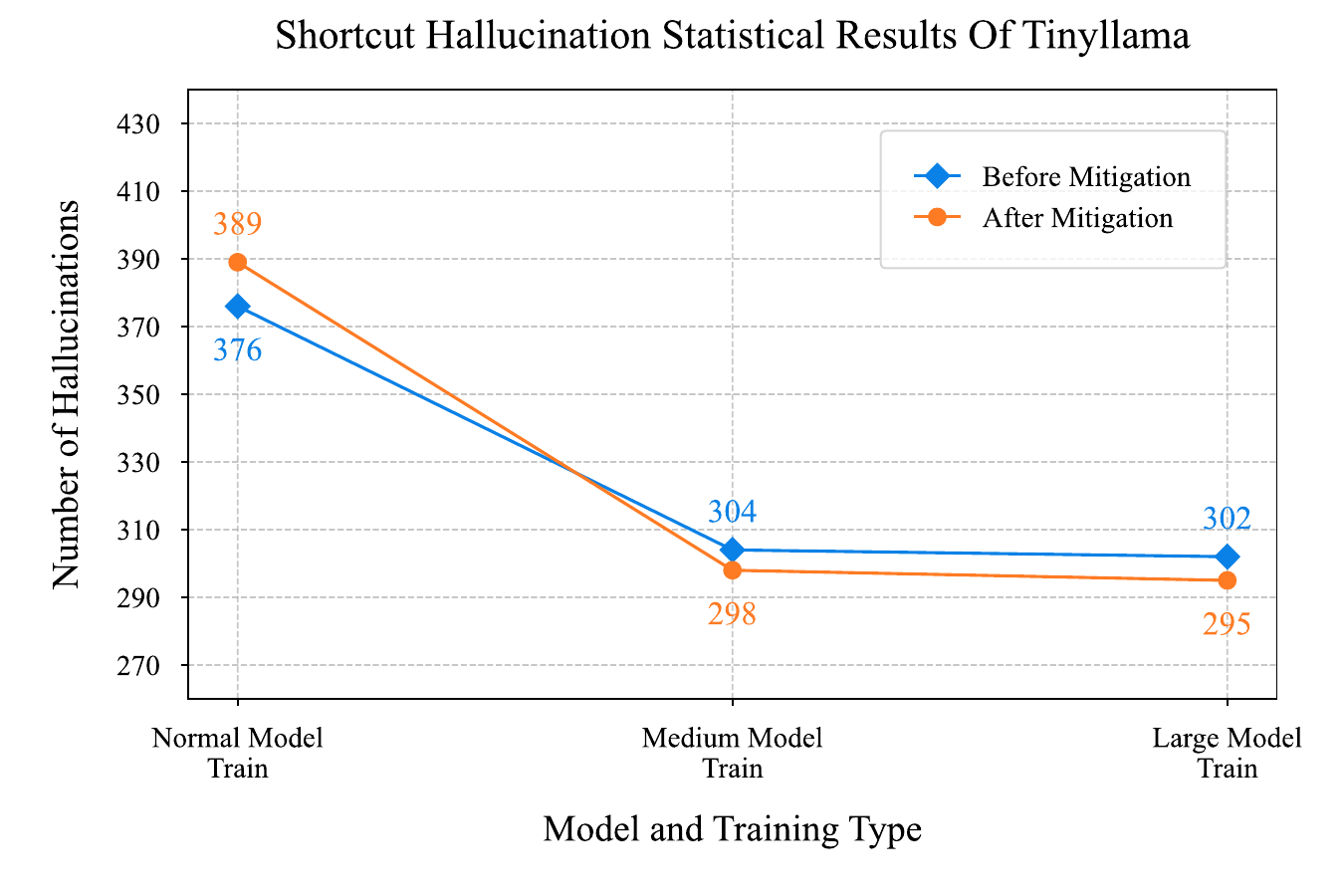}
  \caption {The number of Knowledge-Shortcut hallucination in CQA tasks before and after mitigation in new experiments}
  \label{fig: new number of ksh}
\end{figure*}

\subsection{Reproducibility of Experimental Results}
\label{reproducible repeated experiments}
To minimize the impact of randomness on our experimental results, we conducted multiple repeated experiments and have open-sourced all code and results. The full reproducible experiments can be accessed and executed to obtain the exact experimental outcomes.  

For these repeated experiments, we primarily focused on nanoGPT. We retrained, generated responses, and conducted testing to obtain a comprehensive set of experimental results, which are presented below.

\begin{figure*}[t]
  \includegraphics[width=0.3\linewidth]{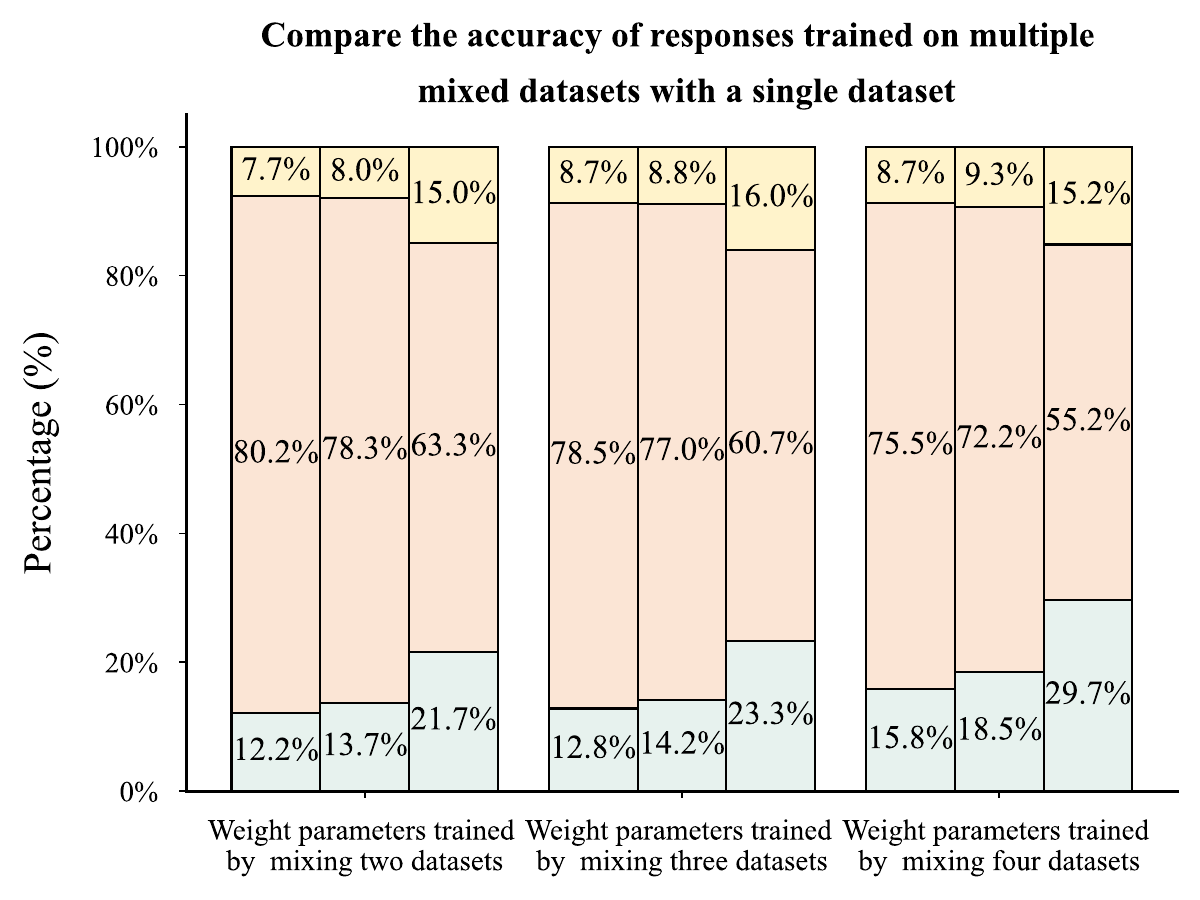} \hfill
  \includegraphics[width=0.3\linewidth]{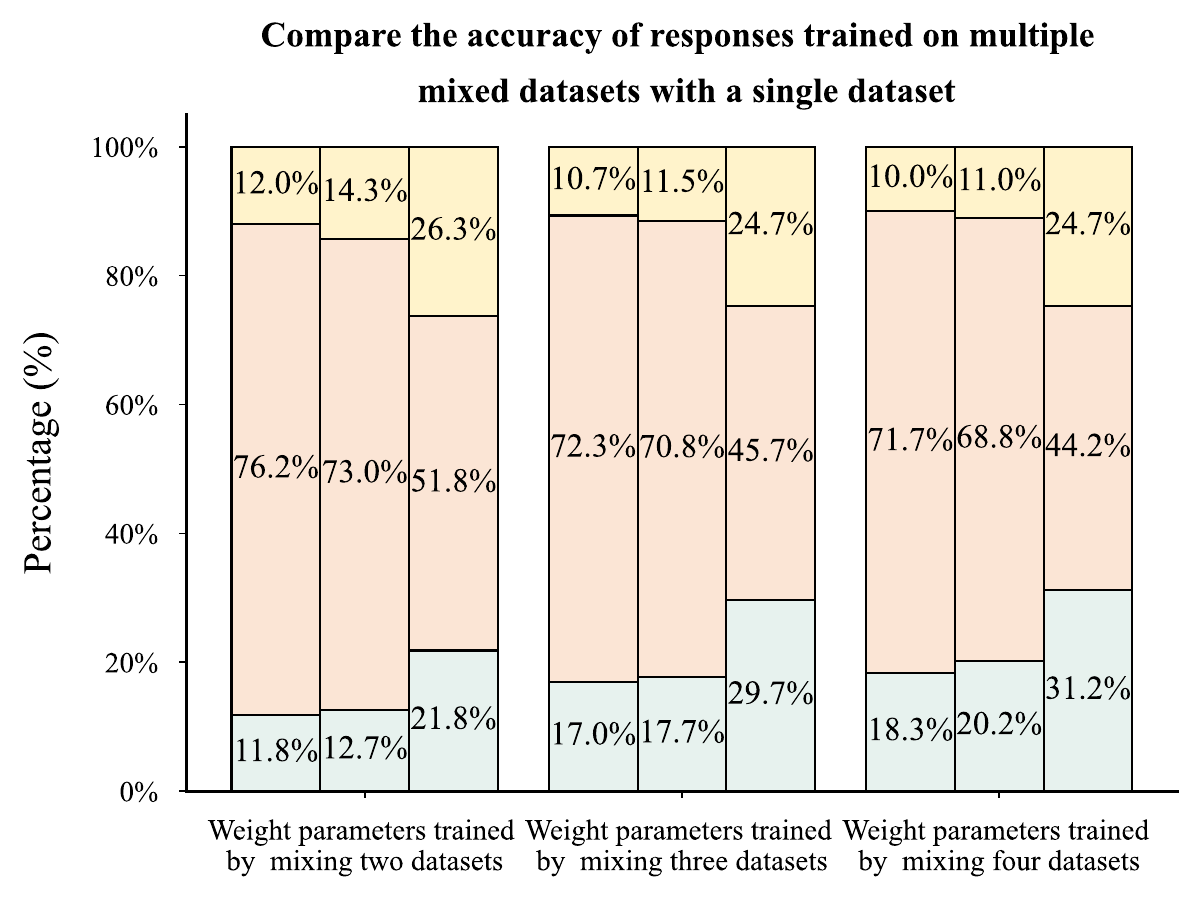} \hfill
  \includegraphics[width=0.36\linewidth]{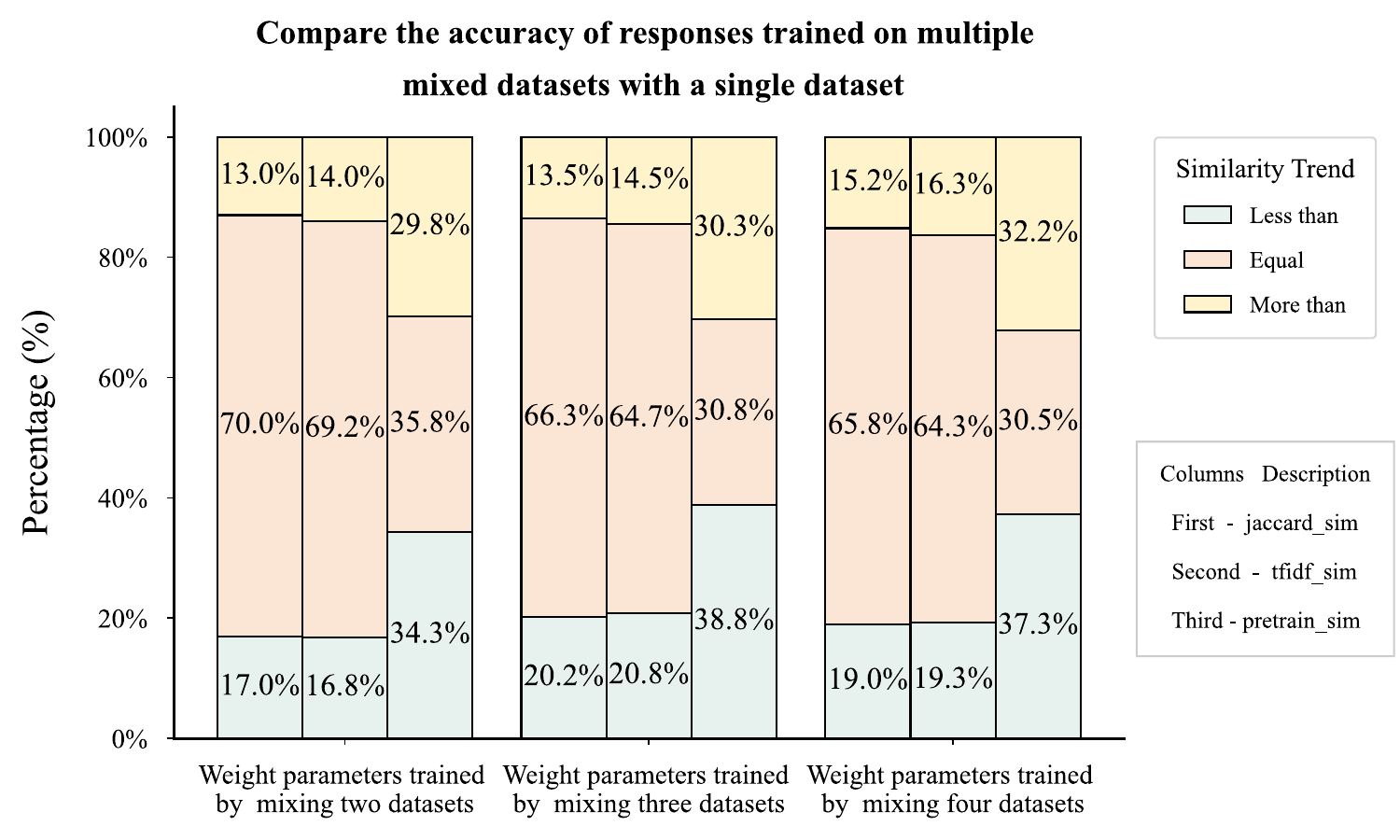}
  \caption {Reproduce experiments, before mitigation, fine-tuning: nanoGPT large, medium, normal}
  \label{fig:new ft-before}
\end{figure*}
\begin{figure*}[t]
  \includegraphics[width=0.3\linewidth]{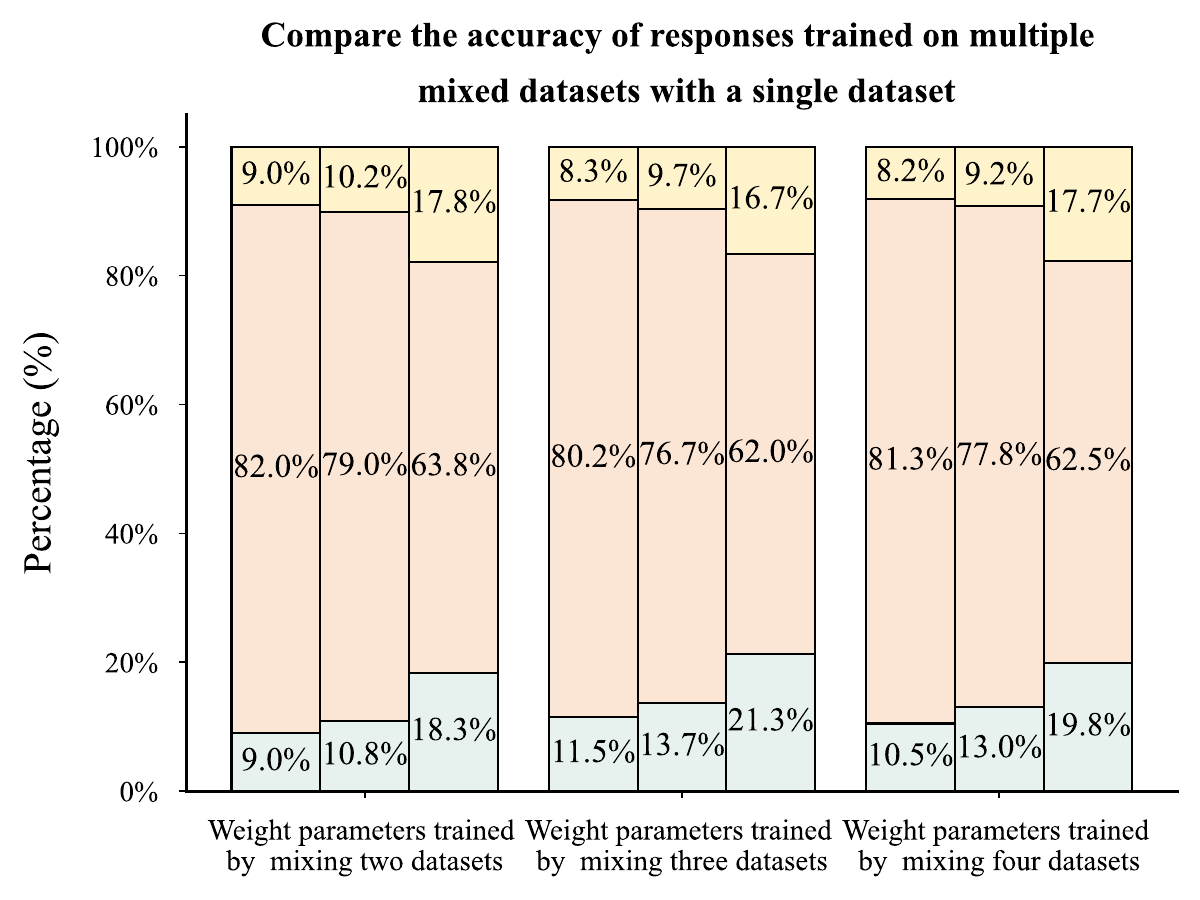} \hfill
  \includegraphics[width=0.3\linewidth]{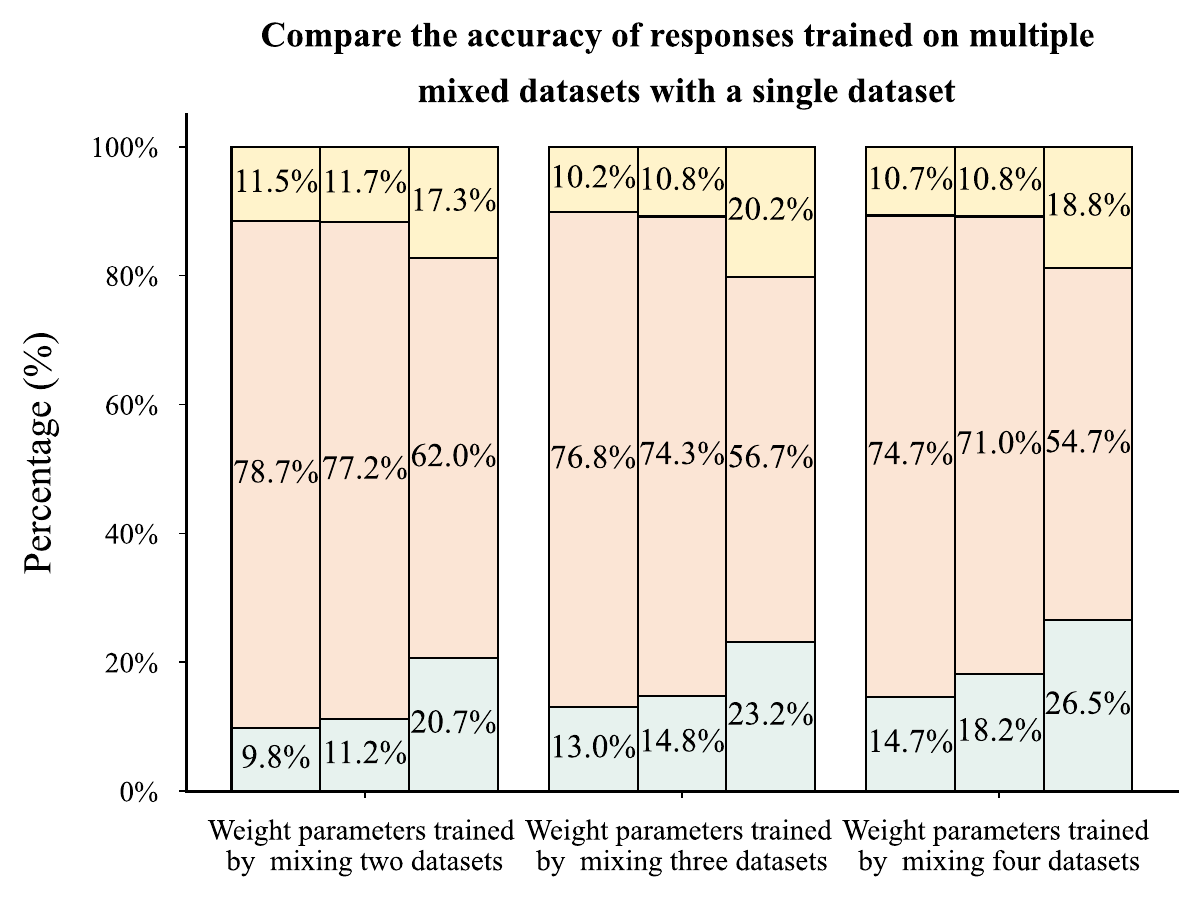} \hfill
  \includegraphics[width=0.36\linewidth]{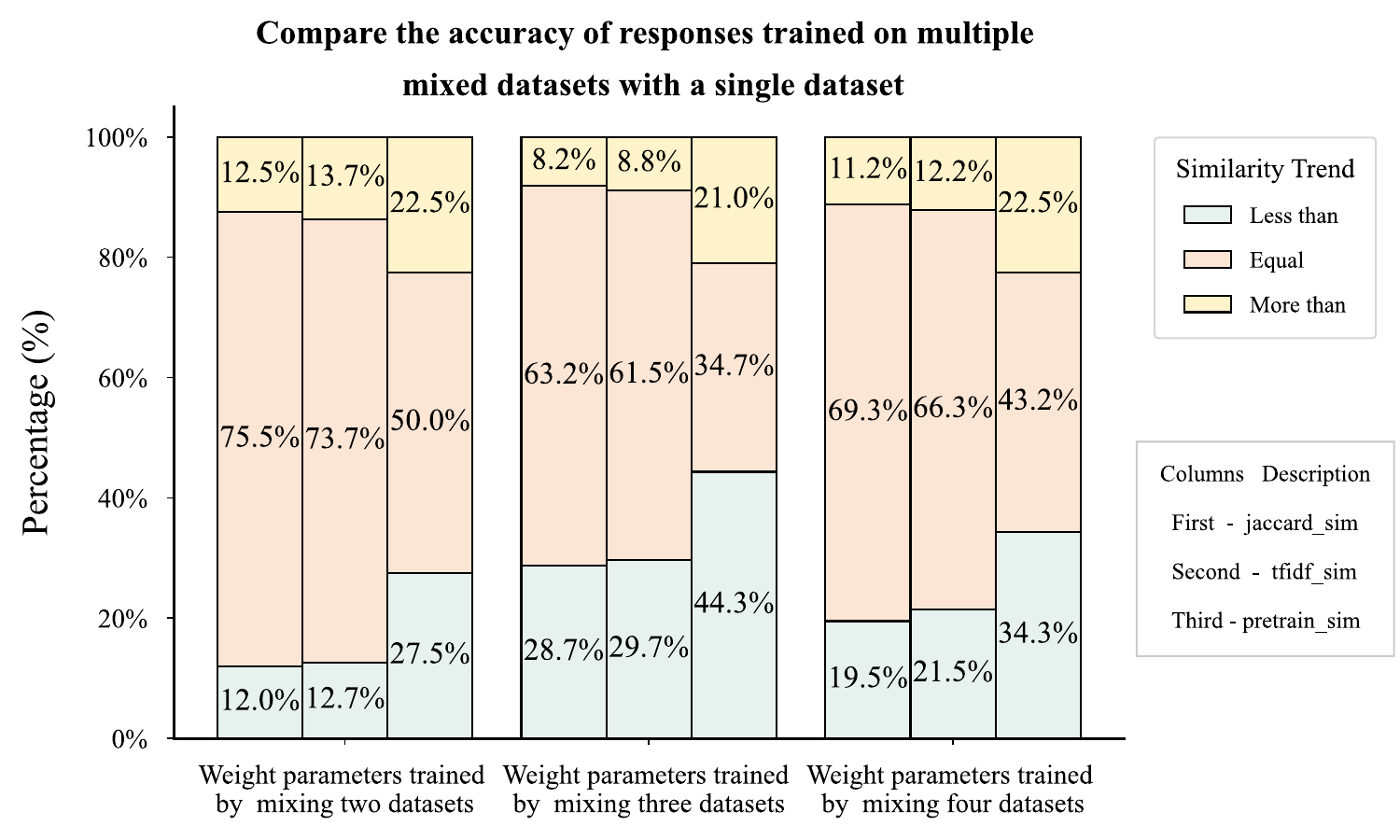}
  \caption {Reproduce experiments, before mitigation, train: nanoGPT large, medium, normal}
  \label{fig:new train-before}
\end{figure*}

\begin{figure*}[t]
  \includegraphics[width=0.3\linewidth]{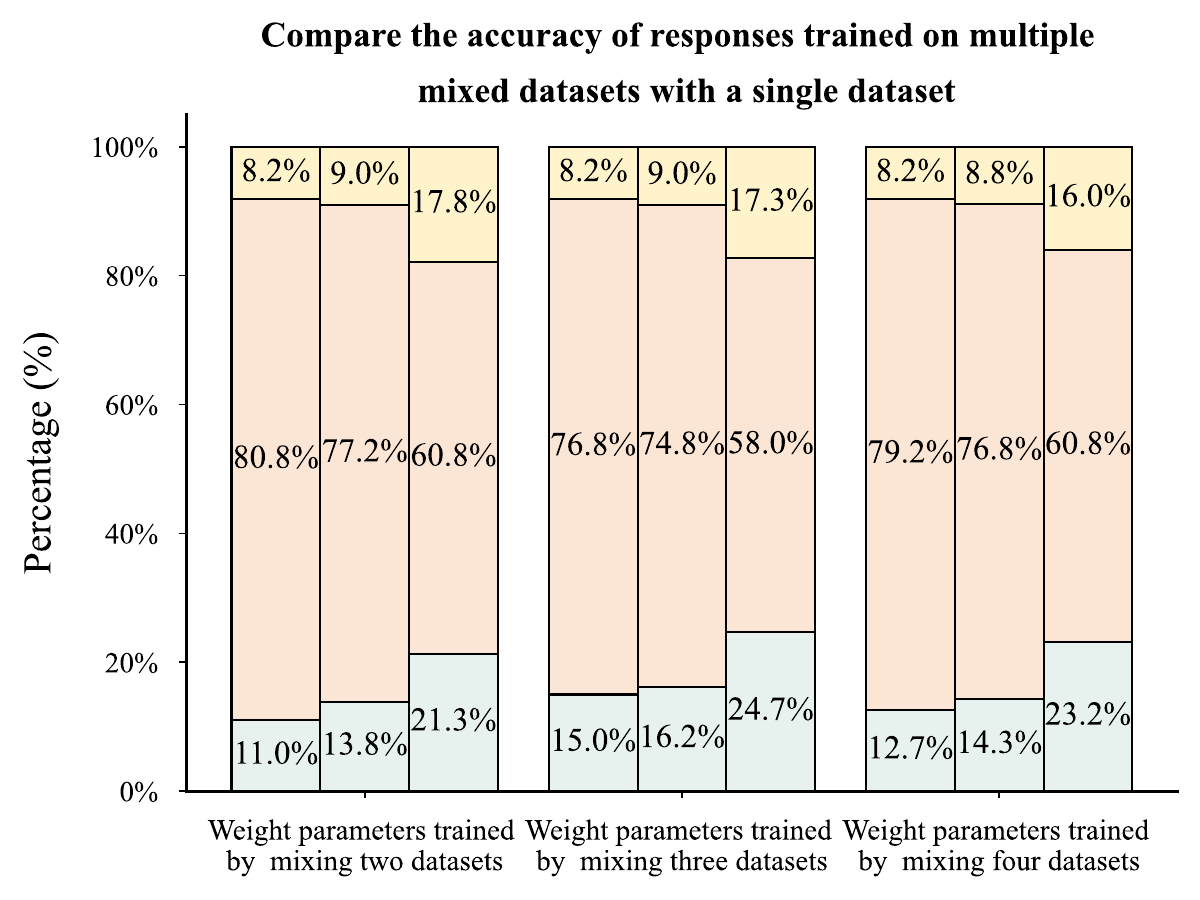} \hfill
  \includegraphics[width=0.3\linewidth]{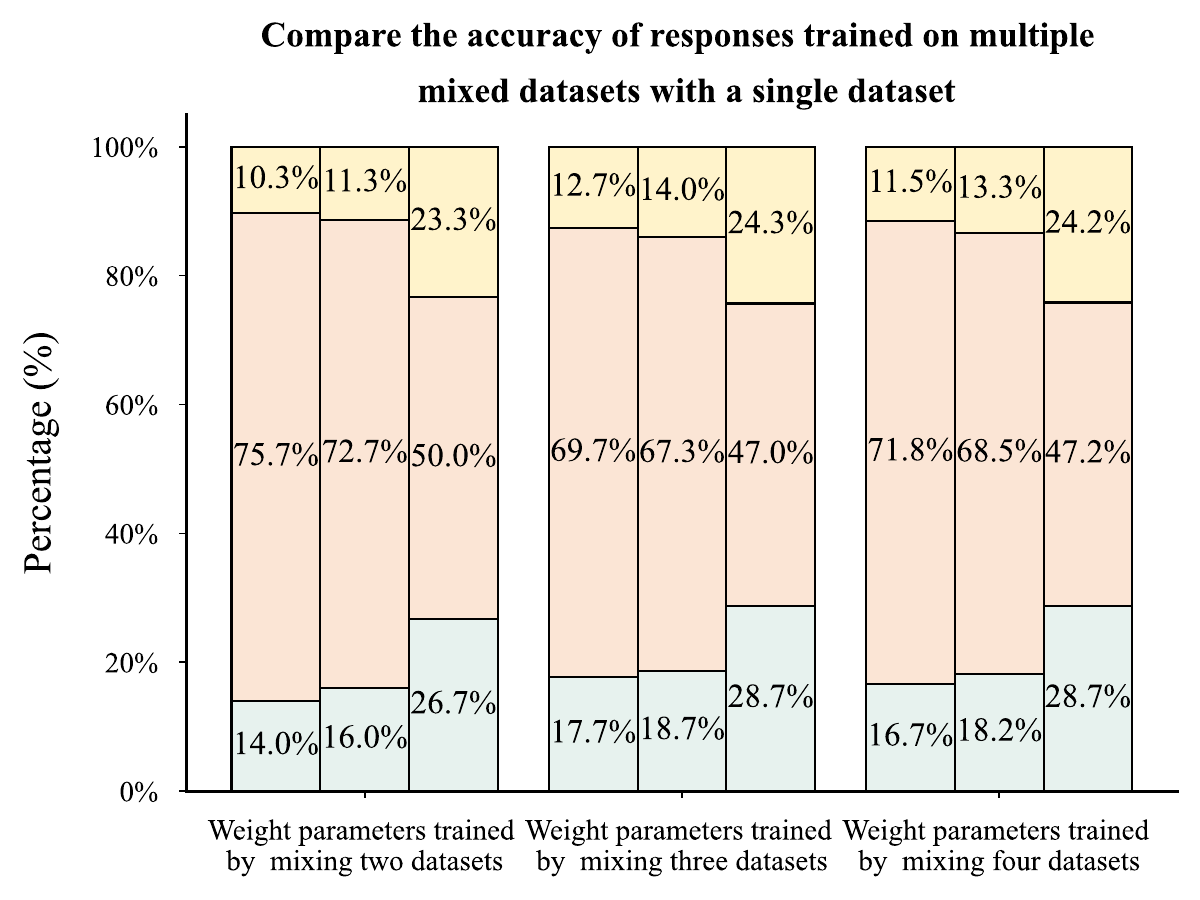} \hfill
  \includegraphics[width=0.36\linewidth]{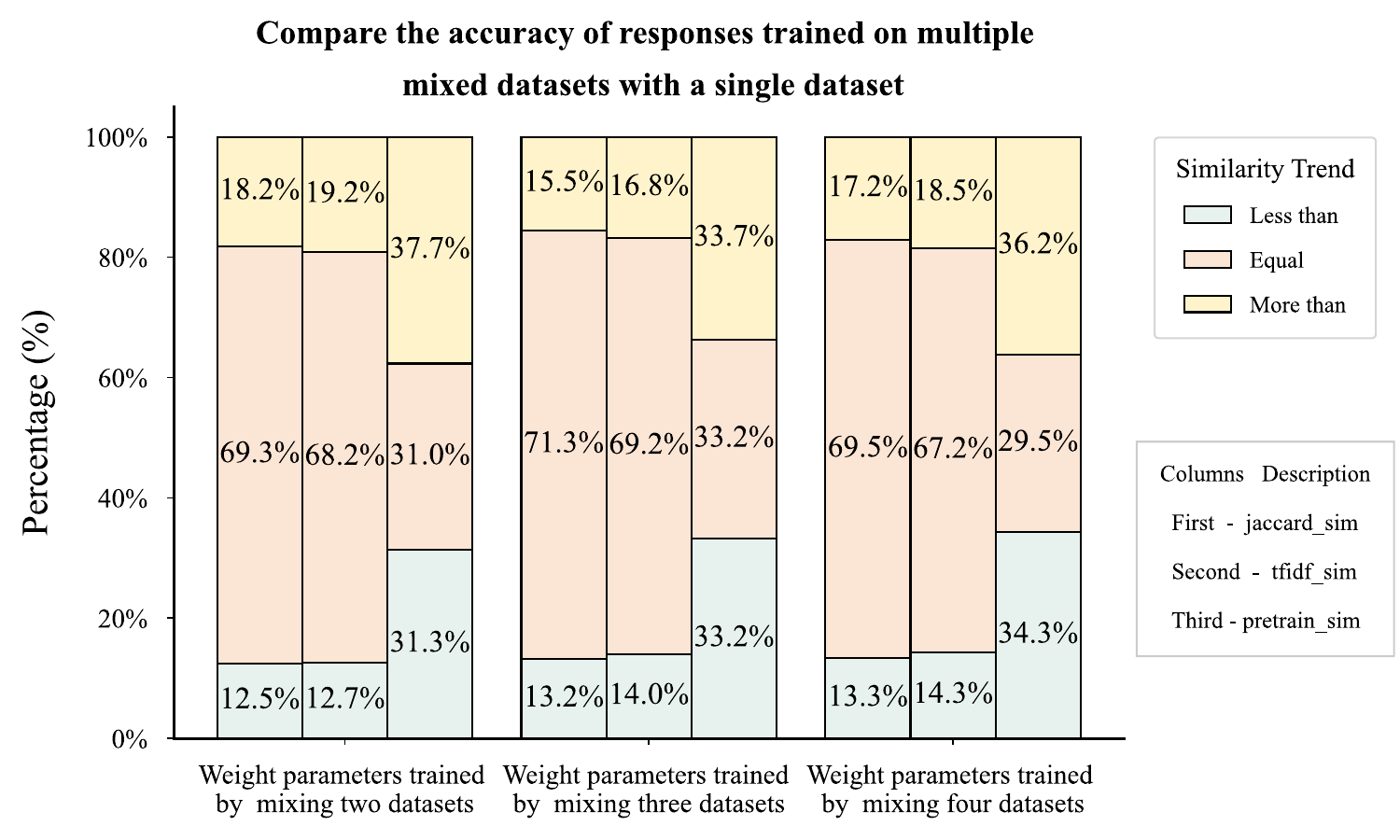}
  \caption {Reproduce experiments, after mitigation, fine-tuning: nanoGPT large, medium, normal}
  \label{fig:new ft-after}
\end{figure*}
\begin{figure*}[t]
  \includegraphics[width=0.3\linewidth]{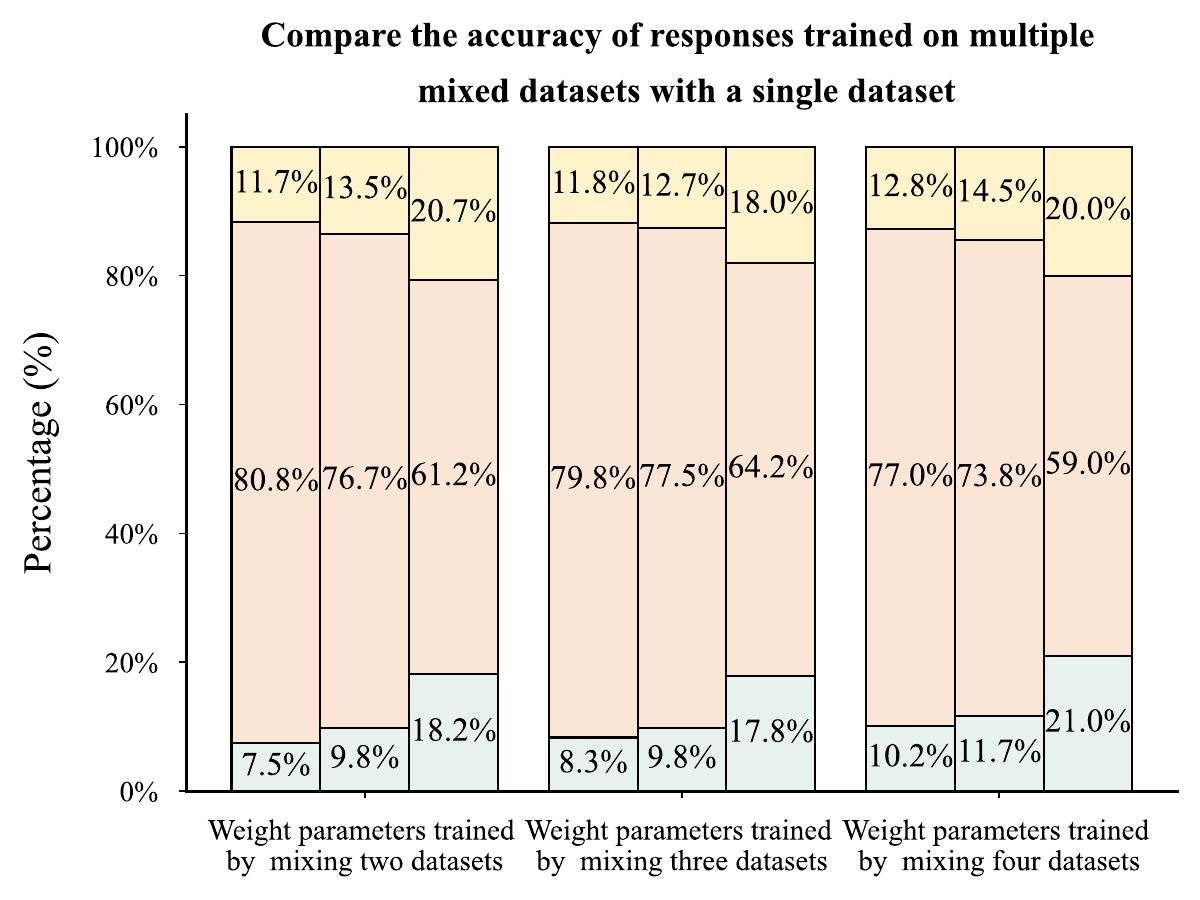} \hfill
  \includegraphics[width=0.3\linewidth]{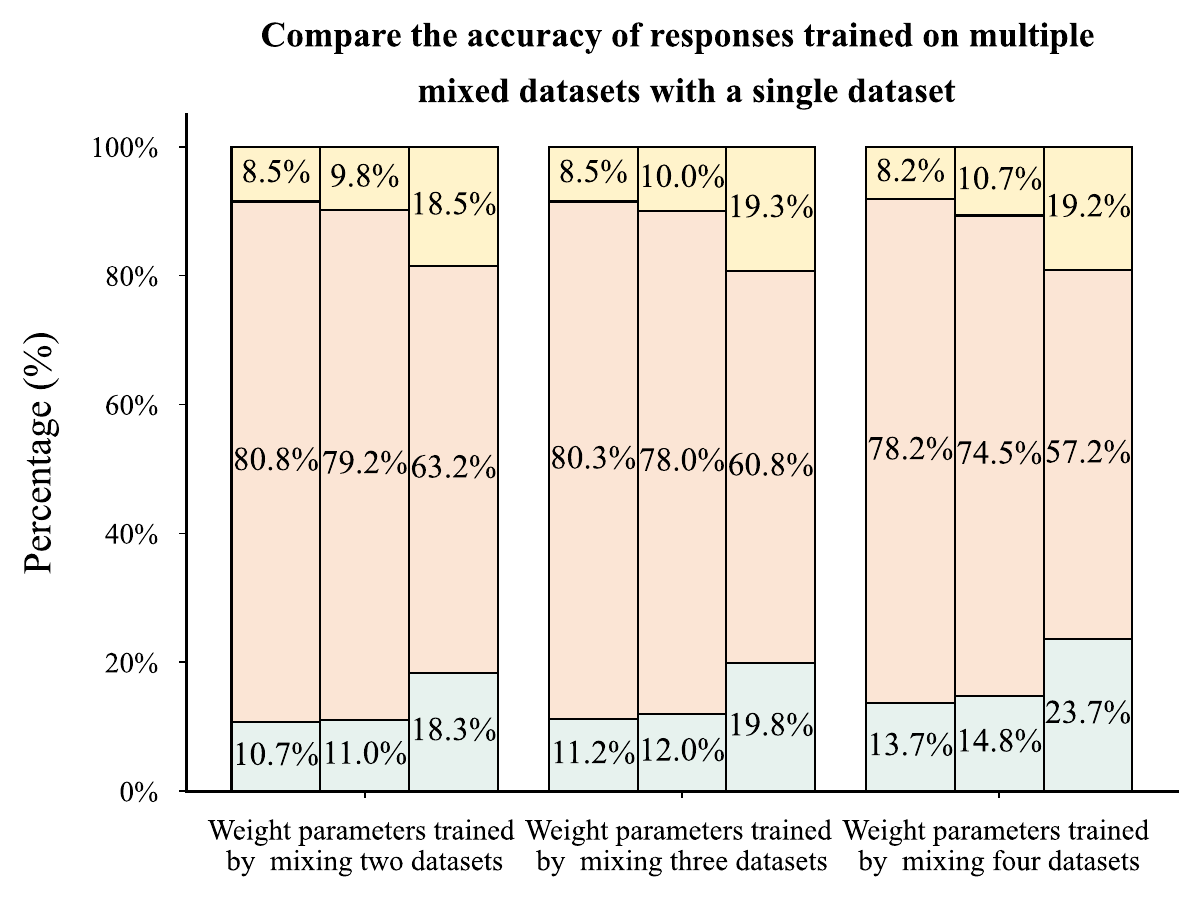} \hfill
  \includegraphics[width=0.36\linewidth]{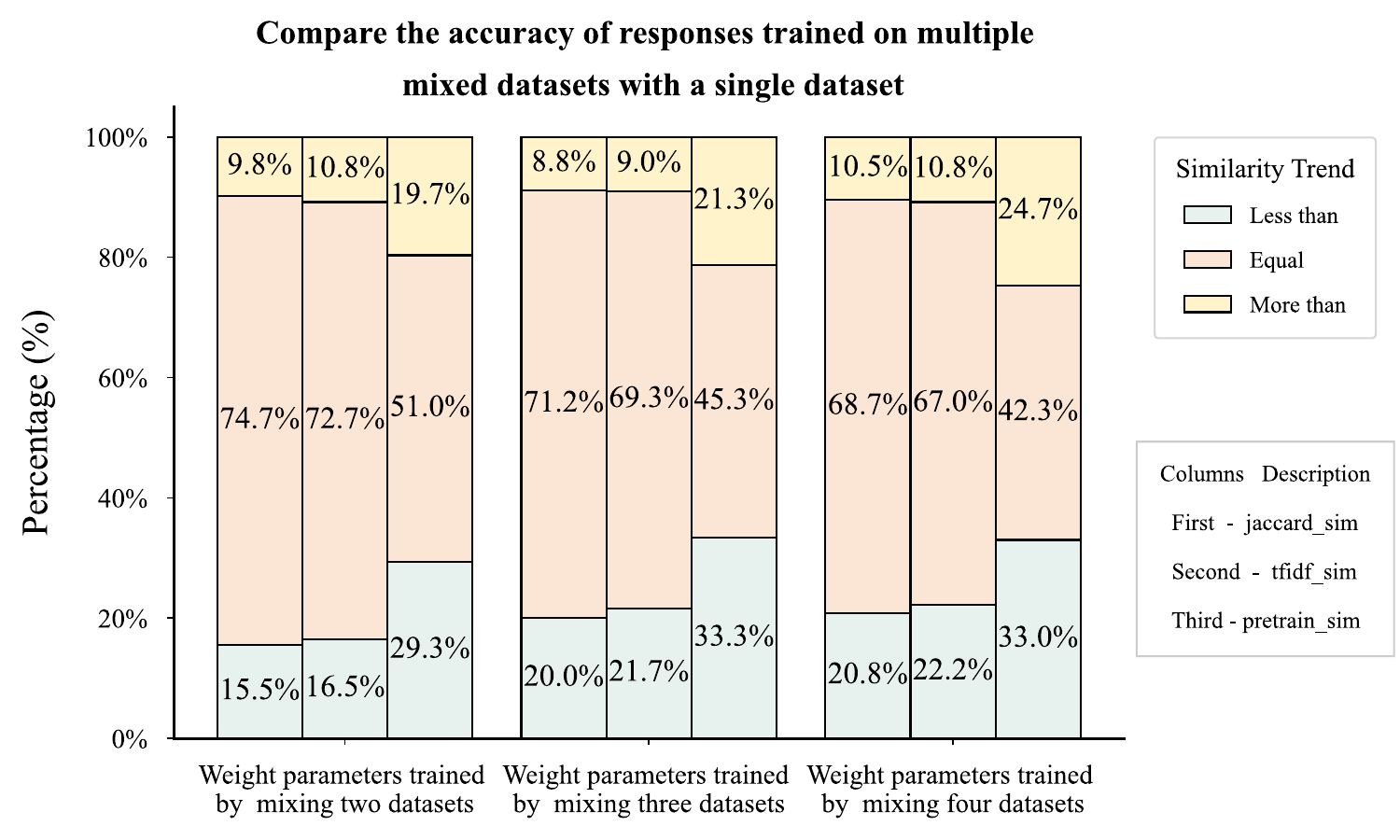}
  \caption {Reproduce experiments, after mitigation, train: nanoGPT large, medium, normal}
  \label{fig:new train-after}
\end{figure*}

\begin{table*}[htbp!]
\centering
\begin{tabular}{llcccccc}
\toprule
\textbf{Type} & \textbf{Parameter} & \multicolumn{3}{c}{\textbf{After mitigation}} & \multicolumn{3}{c}{\textbf{Before mitigation}} \\
\cmidrule(lr){3-5} \cmidrule(lr){6-8}
& & \textbf{Jaccard} & \textbf{TF-IDF} & \textbf{Pre-train} & \textbf{Jaccard} & \textbf{TF-IDF} & \textbf{Pre-train} \\
\midrule
\multirow{6}{*}{\textbf{fine-tune}} 
& large   & \textbf{1638} & \textbf{1647} & 2400 & 1612 & 1618 & 2400 \\
&                & \textbf{0.61965} & \textbf{0.60919} & \textbf{0.76633} & 0.61193 & 0.60121 & 0.76393 \\
& medium  & 1380 & 1390 & 2400 & \textbf{1388} & \textbf{1399} & 2400 \\
&                & 0.50376 & 0.49043 & 0.69370 & \textbf{0.51097} & \textbf{0.50020} & \textbf{0.69581} \\
& normal  & 949 & 967 & 2400 & \textbf{995} & \textbf{1012} & 2400 \\
&                & 0.32107 & 0.31457 & 0.55900 & \textbf{0.34018} & \textbf{0.33170} & \textbf{0.57477} \\
\midrule
\multirow{6}{*}{\textbf{train}} 
& large   & \textbf{1858} & \textbf{1870} & \textbf{2400} & 1844 & 1853 & 2398 \\
&                & \textbf{0.71944} & \textbf{0.70863} & \textbf{0.83367} & 0.71422 & 0.70078 & 0.82887 \\
& medium  & \textbf{1742} & \textbf{1755} & 2400 & 1716 & 1725 & 2400 \\
&                & \textbf{0.66459} & \textbf{0.65435} & \textbf{0.79688} & 0.65474 & 0.64318 & 0.79082 \\
& normal  & \textbf{1370} & \textbf{1382} & 2400 & 1331 & 1346 & 2400 \\
&                & \textbf{0.50535} & \textbf{0.49617} & \textbf{0.68918} & 0.49451 & 0.48467 & 0.67837 \\
\bottomrule
\end{tabular}
\caption{Results with mitigation and before mitigation for various nanoGPT parameters. Values in \textbf{bold} denote significant results.}
\label{tab:new after mitigation evaluate}
\end{table*}

\begin{table*}[htbp]
\centering
\begin{tabular}{lcccp{8cm}} 
\toprule
\textbf{Type} & \textbf{HS Index} & \textbf{Dataset} & \textbf{Row Index} & \textbf{Context} \\ 
\midrule
Jaccard & 17 & 3 & 1326 & Insulin is a hormone made up of a small polypeptide protein that is secreted by the pancreas, which acts as both an endocrine and exocrine \colorbox{red}{gland}. Endocrine \colorbox{red}{glands} are the system of \colorbox{red}{glands} that secrete hormones to regulate body functions. Exocrine \colorbox{red}{glands} aid in digestion. \\
Jaccard & 18 & 3 & 11791 & Epinephrine (ep-uh-nef-rin, -reen) is also known as adrenaline. It is a hormone that is secreted by the \colorbox{red}{adrenal glands}. \\
Jaccard & 49 & 3 & 1303 & The thyroid \colorbox{red}{gland} is one of the body's most important endocrine organs...... \\
\midrule
TF-IDF & 45 & 3 & 11235 & Pinnipeds have streamlined, spindle-shaped bodies with reduced or non-existent external ear flaps, rounded heads, flexible necks, limbs modified into flippers, and small tails...... The mammary \colorbox{red}{glands} and genitals of pinnipeds can retract into the body. \\
\bottomrule
\end{tabular}
\caption{Examples of high-frequency co-occurring words found in the high similarity group}
\label{tab:example for lone context}
\end{table*}

\end{document}